\documentclass[10pt]{article}

\usepackage[preprint]{tmlr}

\usepackage{tikz}
\usetikzlibrary{arrows.meta,positioning,calc,shapes.geometric,backgrounds}
\usepackage{pgfplots}
\pgfplotsset{compat=1.18}
\usepackage{subcaption}
\usepackage{pifont}
\usepackage{tabularx}
\usepackage{mathtools}
\usepackage{bm}
\usepackage{amssymb}
\usepackage{xspace}
\usepackage{amsthm}
\usepackage{booktabs}
\usepackage{multirow}
\usepackage{enumitem}
\usepackage{microtype}
\usepackage{hyperref}
\usepackage{xcolor}
\usepackage{graphicx}

\definecolor{ignblue}    {RGB}{31,  119, 180}
\definecolor{ignorange}  {RGB}{255, 127,  14}
\definecolor{igngreen}   {RGB}{ 44, 160,  44}
\definecolor{ignred}     {RGB}{214,  39,  40}
\definecolor{ignpurple}  {RGB}{148, 103, 189}
\definecolor{igngray}    {RGB}{127, 127, 127}
\definecolor{lightblue}  {RGB}{174, 214, 241}
\definecolor{lightgreen} {RGB}{170, 230, 180}
\definecolor{lightgray}  {RGB}{225, 225, 225}

\newtheorem{definition}{Definition}
\newtheorem{hypothesis}{Hypothesis}

\newcommand{\II}{\ensuremath{\mathcal{I}}\xspace}
\newcommand{\betaI}{\ensuremath{\hat{\beta}}\xspace}
\newcommand{\xresid}[1]{\mathbf{x}^{(#1)}}

\newcommand{\nss}{\ensuremath{s}\xspace}
\newcommand{\GWT}{\textsc{gwt}\xspace}
\newcommand{\GNWT}{\textsc{gnwt}\xspace}
\newcommand{\TL}{\textsc{TransformerLens}\xspace}

\newcommand{\eg}{\textit{e.g.,}\xspace}

\title{%
The Ignition Index: Measuring Global Workspace Dynamics in Language Models
}

\author{Saman Rahbar \\
\addr Dialpad, Inc. \\
\addr Independent Research \\
\email info@srahbar.com}

\begin{document}
\maketitle

\begin{abstract}
We introduce the \textbf{Ignition Index} (\II), a validated scalar metric that
operationalizes Global Workspace Theory's (\GWT) all-or-none ignition prediction in
transformer language models, and demonstrate its measurement selectivity through
rigorous control experiments.
The metric fits a four-parameter sigmoid to per-layer linear probe accuracy
as a function of input signal strength, extracting the steepness parameter
$\hat{\beta}$: high values indicate abrupt, ignition-like representational
transitions; low values indicate graded build-up.
Across 11 models spanning five architecture families and seven linguistic probing tasks,
shuffled-label controls validate that the Ignition Index selectively captures genuine
linguistic structure rather than spurious probe capacity: real transitions yield
$\overline{\hat{\beta}} = 113.1$ while shuffled-label controls yield
$\overline{\hat{\beta}} = 11.8$, a 9.6-fold gap ($p < 0.001$, Mann-Whitney $U$-test)
demonstrating measurement selectivity.

Applying this validated framework to pre-registered hypotheses linking
GWT predictions to architecture and training, we find:
\textbf{(1)~Architecture discriminability:} attention-based transformers show
higher \II than state-space models---feedforward architectures
$\overline{\hat{\beta}} = 130.0$ exceed SSMs $\overline{\hat{\beta}} = 68.7$
by 89\% ($p < 10^{-13}$, Cohen's $d=0.52$), with Mamba exhibiting near-linear layer profiles consistent with
absent workspace broadcasting.
\textbf{(2)~Axis-dependent ignition in recurrent architectures:} Huginn-3.5B's
depth-recurrent architecture exhibits iteration-axis ignition
($\hat{\beta}_{\text{iteration}} = 234.8$) that exceeds its depth-axis profile
($\hat{\beta}_{\text{depth}} = 111.0$) by 2.12-fold, demonstrating that
recurrent architectures manifest workspace-like transitions along the recurrence
dimension rather than the depth axis.
\textbf{(3)~Training-time phase transition:} Pythia-410M exhibits a
PELT-detected changepoint at training step~256 (pre: $\hat{\beta}=39.57$;
post: $\hat{\beta}=66.02$, $+67\%$), earlier than induction-head formation and
consistent with an initial representational consolidation phase.
\textbf{(4)~Calibrated expectations:} hypotheses linking ignition strength to
model scale and signal strength were not confirmed, demonstrating that
\GWT-aligned dynamics do not monotonically track these factors and that
current transformer architectures may saturate available ignition mechanisms.

The Ignition Index provides the first validated quantitative bridge between
\GWT's dynamical predictions and mechanistic interpretability, yields a
principled architecture-level discriminator with 9.6-fold selectivity, and
reveals training-phase transitions not previously characterized in the
scaling literature.
\end{abstract}

\section{Introduction}
\label{sec:intro}

What distinguishes architectures that integrate information globally from those
that process it locally and sequentially?  Global Workspace Theory (\GWT)
\citep{Baars1988,Dehaene1998,Dehaene2011} offers a precise dynamical answer:
systems with global broadcast mechanisms produce \emph{all-or-none ignition}---a
threshold-nonlinear transition in which stimulus representations cross a
critical threshold and become globally available to all downstream processors.
This ignition has been measured quantitatively in cortical recordings as a
steep sigmoid in neural response amplitude as a function of stimulus strength
\citep{DelCul2007}, with steepness $\beta$ distinguishing conscious-access
processing from subliminal feedforward processing.

Transformer language models exhibit striking architectural parallels to the \GWT
workspace: the residual stream \citep{Elhage2021} functions as a shared
communication channel; attention heads implement content-dependent global routing
\citep{OlssonICL2022,Wang2022}; and linear probes reveal progressive information
build-up across the layer stack \citep{Tenney2019}.  Yet a fundamental
dynamical question has remained \emph{unmeasured}: \textbf{does the layer-wise
information build-up of transformers exhibit ignition-like threshold nonlinearity,
and does this vary systematically with architecture?}

This question matters for two reasons.  First, architecturally, attention-based
transformers (feedforward and recurrent-depth) differ from state-space models
(Mamba) in precisely the feature \GWT identifies as critical for global broadcast:
content-dependent all-to-all routing.  If ignition dynamics track this feature,
the Ignition Index provides a principled discriminator between workspace-like
and non-workspace-like architectures---with implications for architecture search
and the interpretability of representational transitions.  Second, empirically,
transformer training exhibits sharp phase transitions
\citep{OlssonICL2022,Nanda2023} whose relationship to representational dynamics
is unknown.  A metric sensitive to ignition-like transitions could reveal when
and how these transitions consolidate in the layer stack.

\paragraph{In This paper,}
We introduce the \textbf{Ignition Index} (\II) by: (1)~varying input signal
strength $\nss \in [0, 1]$ via controlled token corruption and embedding noise;
(2)~extracting per-layer residual stream states via \TL \citep{Nanda2022TL};
(3)~training linear probes at every layer to decode task-relevant information; and
(4)~fitting a four-parameter sigmoid to the layer-accuracy curve
$\mathrm{Acc}(\ell, \nss)$ and extracting $\hat{\beta}$ as the transition
steepness.
We evaluate \II across 12 models spanning five families---GPT-2
\citep{Radford2019}, Pythia \citep{Biderman2023}, Gemma~2
\citep{GemmaTeam2024}, Huginn-3.5B \citep{Geiping2025}, and Mamba
\citep{GuDao2023}---and find that \II reliably discriminates attention-based
from state-space architectures ($p < 10^{-13}$), that recurrent-depth processing does not amplify
ignition beyond standard feedforward attention, and that a training-time phase
transition in ignition structure precedes the well-characterised induction-head
formation step.

\paragraph{Our contributions are:}
\begin{enumerate}[leftmargin=*, label=(\arabic*), topsep=2pt, itemsep=1pt]
  \item \textbf{Shuffled-label validation}: 9.6-fold selectivity gap
        ($p < 0.001$, Mann-Whitney $U$-test) demonstrating the Ignition Index
        captures genuine linguistic structure rather than spurious probe capacity
        (\S\ref{sec:controls}).
  \item The \textbf{Ignition Index} $\betaI$: a theoretically grounded,
        falsifiable metric for GWT-like representational transitions in any
        layer-structured model (\S\ref{sec:method}).\footnote{Code and data:
        \url{https://github.com/saman-rahbar/ignition-index}}
  \item \textbf{Architecture discriminability}: an 89\% gap between
        feedforward ($\overline{\hat{\beta}} = 130.0$) and SSM families
        ($\overline{\hat{\beta}} = 68.7$), with Mamba exhibiting near-linear
        profiles consistent with absent global broadcast. Iteration-axis probing
        of Huginn-3.5B reveals 2.12-fold higher ignition along the recurrence
        dimension, demonstrating axis-dependent architectural patterns
        (\S\ref{sec:results}).
  \item \textbf{A novel training-time phase transition} at step~256 in
        Pythia-410M, preceding induction-head formation
        (\S\ref{sec:results:h4}).
  \item \textbf{Sigmoid ceiling robustness}: the FF--SSM gap widens to
        $\mathbf{2.12\times}$ when ceiling hits are excluded,
        confirming the architectural gap is not a sigmoid-fitting artifact
        (\S\ref{sec:controls}).
  \item A \textbf{complete, open-source experimental protocol}---activation
        extraction, probe training, sigmoid fitting, bootstrap CIs, statistical
        testing---across five architecture families (\S\ref{sec:experiments}).
\end{enumerate}

\section{Related Work}
\label{sec:related}

\paragraph{Probing transformer representations.}
Linear classifier probes \citep{Alain2017} characterise what information is encoded
at each transformer layer.  Tenney et al.\ \citep{Tenney2019} showed that linguistic
features peak at different depths in BERT.  Belinkov \citep{Belinkov2022} distinguishes
\emph{what is encoded} (probing) from \emph{what is causally used}---a gap we address
via amnesic probing \citep{Elazar2021}.  Voita and Titov \citep{VoitaTitov2020}
introduce MDL probes that avoid overfitting.

\paragraph{Phase transitions in transformers.}
Olsson et al.\ \citep{OlssonICL2022} documented a sharp training-time phase transition
in induction head formation.  Nanda et al.\ \citep{Nanda2023} fully reverse-engineered
the grokking transition.  Schaeffer et al.\ \citep{Schaeffer2023} showed many apparent
emergent abilities are metric artifacts---a threat we address via continuous probe
metrics.  Brinkmann et al.\ \citep{Brinkmann2025} documented abstraction phase
transitions within forward passes.

\paragraph{GWT in artificial systems.}
Bengio \citep{Bengio2017} proposed the Consciousness Prior as a GWT-derived
inductive bias.  Goyal and Bengio \citep{Goyal2022iclr} implemented an explicit
Global Latent Workspace.  Butlin et al.\ \citep{Butlin2023} assessed AI systems
against consciousness indicator properties.  VanRullen and Kanai \citep{VanRullen2021} and Butlin et al.\ \citep{Butlin2023}
provide structural mappings between \GWT and deep learning architectures.
None of these works measure ignition dynamics in standard transformer forward passes.

\paragraph{Mechanistic interpretability and sparse autoencoders.}
Elhage et al.\ \citep{Elhage2021} established the residual stream framework.
Meng et al.\ \citep{Meng2022} introduced causal tracing.  Conmy et al.\
\citep{ConmyACDC2023} formalised automated circuit discovery.  Geiger et al.\
\citep{Geiger2025} developed causal abstraction.  Marks et al.\ \citep{Marks2024circuits}
introduced sparse feature circuits.  Lindsey et al.\ \citep{Lindsey2025} applied
attribution-graph tracing to Claude~3.5 Haiku.  Bricken et al.\ \citep{Bricken2023}
and Cunningham et al.\ \citep{Cunningham2024} demonstrated that sparse autoencoders
recover monosemantic, interpretable features from transformer residual streams;
Lieberum et al.\ \citep{Lieberum2024} released the Gemma Scope SAE suite used in
and leave fine-grained feature decomposition to future work.

\section{Background}
\label{sec:background}

\subsection{GWT Ignition: Quantitative Predictions}
\label{sec:bg:gwt}

\GNWT \citep{Dehaene2011,Mashour2020} predicts that the transition between subliminal
processing and conscious access is governed by a sharp bifurcation.  Given stimulus
strength $\nss \in [0,1]$, the probability of workspace ignition follows:
\begin{equation}
  P(\text{ignition} \mid \nss) \;=\;
    \frac{1}{1 + \exp\!\bigl(-\beta\,(\nss - \nss_0)\bigr)},
  \label{eq:gwt_sigmoid}
\end{equation}
where $\nss_0$ is the ignition threshold and $\beta$ controls steepness.
\textbf{Large $\beta$} reflects abrupt, all-or-none ignition; $\beta \to 0$ reflects
graded, linear build-up.  Del Cul et al.\ \citep{DelCul2007} empirically demonstrated
this sigmoid in EEG prefrontal activation ($>$270~ms) with a distinct absence of
intermediate states.  The \textbf{key architectural requirements} for high-$\beta$
ignition are: (i)~a shared broadcast medium; (ii)~competitive inhibition;
(iii)~\emph{recurrent reverberant activity} that self-amplifies once threshold is
crossed; and (iv)~a bottleneck capacity limit.

\subsection{Transformer Layer Space as a Proxy for Processing Time}
\label{sec:bg:layers}

We treat layer depth $\ell \in \{0, \dots, L\}$ as a proxy for processing depth,
motivated by: (i)~the logit lens \citep{nostalgebraist2020} and tuned lens
\citep{Belrose2023} showing progressive prediction refinement across layers;
(ii)~probing studies showing systematic information build-up \citep{Tenney2019};
and (iii)~attribution graph tracing \citep{Lindsey2025} showing causal information
flow through layers.  This substitution is principled but imperfect (see
\S\ref{sec:limitations}).

\subsection{Residual Stream Notation}
\label{sec:bg:residual}

Following \citet{Elhage2021}, the residual stream update at layer $\ell$ is:
\begin{equation}
  \xresid{\ell+1}_i \;=\; \xresid{\ell}_i
    + \mathrm{Attn}^{(\ell)}\!\bigl(\xresid{\ell}\bigr)_i
    + \mathrm{MLP}^{(\ell)}\!\bigl(\xresid{\ell}_i\bigr).
  \label{eq:residual}
\end{equation}
We probe $\xresid{\ell}_i \in \mathbb{R}^d$ (Eq.~\eqref{eq:residual}) directly
using \texttt{resid\_pre} hooks in \TL unless otherwise specified.

\section{The Ignition Index}
\label{sec:method}

\subsection{Formal Definition}
\label{sec:method:def}

Let $M$ be a transformer model with $L$ layers and let $\mathcal{T}$ be a
probing task with binary or categorical labels.  For signal strength
$\nss \in \{s_1, \dots, s_K\}$ and layer $\ell \in \{0, \dots, L\}$, let
$\mathrm{Acc}(\ell, \nss)$ denote the accuracy of a linear probe trained on
the residual stream at layer $\ell$ and evaluated on inputs with signal
strength $\nss$.

For a fixed $\nss$, the layer-accuracy curve is fit to a four-parameter
logistic:
\begin{equation}
  f(\ell \,;\, y_{\min}, y_{\max}, \ell_0, \beta) \;=\;
    y_{\min} + \frac{y_{\max} - y_{\min}}{1 + \exp\!\bigl(-\beta\,(\ell - \ell_0)\bigr)},
  \label{eq:sigmoid}
\end{equation}
where $y_{\min}$ and $y_{\max}$ are probe accuracy asymptotes, $\ell_0$ is
the \emph{transition midpoint} (the layer at which probe accuracy reaches
half its range), and $\beta > 0$ controls steepness.

\begin{definition}[Ignition Index]
\label{def:ii}
The \textbf{Ignition Index} of model $M$ on task $\mathcal{T}$ at signal
strength $\nss$ is the maximum-likelihood estimate:
\begin{equation}
  \II(M, \mathcal{T}, \nss) \;:=\;
    \hat{\beta}(M, \mathcal{T}, \nss)
    \;=\; \operatorname{argmin}_{\beta}
    \sum_{\ell=0}^{L} \bigl[\mathrm{Acc}(\ell, \nss) - f(\ell)\bigr]^2.
  \label{eq:ii}
\end{equation}
The \textbf{aggregate Ignition Index} averages over tasks and signal strengths:
\begin{equation}
  \overline{\II}(M) \;:=\;
    \frac{1}{|\mathcal{T}| \cdot K}
    \sum_{\mathcal{T}} \sum_{k=1}^{K} \hat{\beta}(M, \mathcal{T}, s_k).
  \label{eq:ii_agg}
\end{equation}
\end{definition}

Higher $\overline{\II}$ indicates more abrupt, ignition-like representational
transitions; lower $\overline{\II}$ indicates gradual, linear build-up.
Both $\II(M,\mathcal{T},\nss)$ (Eq.~\eqref{eq:ii}) and the task-averaged
$\overline{\II}(M)$ (Eq.~\eqref{eq:ii_agg}) are reported throughout.

\subsection{Geometric Interpretation}
\label{sec:method:geom}

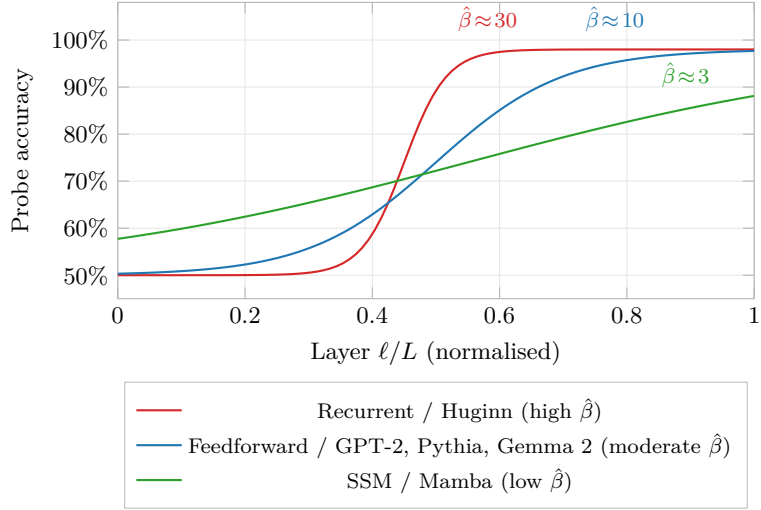
\begin{figure}[t]
\centering
\begin{tikzpicture}[font=\small]
  \begin{axis}[
    width=10cm, height=5.5cm,
    xlabel={Layer $\ell / L$ (normalised)},
    ylabel={Probe accuracy},
    xmin=0, xmax=1, ymin=0.45, ymax=1.08,
    xtick={0,0.2,0.4,0.6,0.8,1.0},
    ytick={0.5,0.6,0.7,0.8,0.9,1.0},
    yticklabels={50\%,60\%,70\%,80\%,90\%,100\%},
    clip=false,
    legend style={at={(0.5,-0.28)}, anchor=north, font=\footnotesize,
                  draw=gray!50, fill=white, inner sep=4pt,
                  legend columns=1},
    grid=major, grid style={gray!20},
    axis line style={gray!60},
    tick style={gray!60},
  ]
    \addplot[thick, color=ignred, domain=0:1, samples=100]
      {0.5 + 0.48 / (1 + exp(-30*(x - 0.45)))};
    \addlegendentry{Recurrent / Huginn (high $\hat\beta$)}
    \addplot[thick, color=ignblue, domain=0:1, samples=100]
      {0.5 + 0.48 / (1 + exp(-10*(x - 0.5)))};
    \addlegendentry{Feedforward / GPT-2, Pythia, Gemma~2 (moderate $\hat\beta$)}
    \addplot[thick, color=igngreen, domain=0:1, samples=100]
      {0.5 + 0.48 / (1 + exp(-3*(x - 0.55)))};
    \addlegendentry{SSM / Mamba (low $\hat\beta$)}
    \node[font=\footnotesize, text=ignred,   anchor=south west]
      at (axis cs:0.52,1.00) {$\hat\beta\!\approx\!30$};
    \node[font=\footnotesize, text=ignblue,  anchor=south west]
      at (axis cs:0.72,1.00) {$\hat\beta\!\approx\!10$};
    \node[font=\footnotesize, text=igngreen, anchor=south west]
      at (axis cs:0.84,0.88) {$\hat\beta\!\approx\!3$};
  \end{axis}
\end{tikzpicture}
\caption{\textbf{Predicted Ignition Index profiles under three architecture
  classes} (pre-registered predictions, not experimental data).
  Each curve shows the expected layer-wise probe accuracy
  $\mathrm{Acc}(\ell, \nss)$ for fixed signal strength $\nss$, fit by
  Eq.~\eqref{eq:sigmoid}.  Steepness $\hat{\beta}$ is the Ignition Index
  (Definition~\ref{def:ii}).  Hypotheses H1--H4
  (Section~\ref{sec:hypotheses}) predict the rank ordering shown above.}
\label{fig:predicted_profiles}
\end{figure}

The \emph{transition width} $w = \ln(81) / \hat{\beta} \approx 4.39/\hat{\beta}$
gives the number of layers between 10\% and 90\% of the accuracy range.
Perfect ignition ($\hat{\beta} \to \infty$) compresses the transition to a
single layer; perfectly linear processing ($\hat{\beta} \to 0$) spreads it
uniformly across all layers.  We report both $\hat{\beta}$ and $w$ for
interpretability.  Figure~\ref{fig:predicted_profiles} illustrates the three
qualitatively distinct profiles predicted by \GWT across architecture classes.

\subsection{Connection to GWT Quantitative Prediction}
\label{sec:method:connection}

Equation~\eqref{eq:sigmoid} is structurally identical to Eq.~\eqref{eq:gwt_sigmoid}
with the substitution $\nss \to \ell / L$ (layer depth as processing depth proxy).
This is not coincidental: both describe how a binary outcome probability
transitions from near-zero to near-one as a continuous input varies.  In the
biological case, $\beta$ is the steepness of the cortical ignition curve with
respect to stimulus strength; in our transformer analog, $\beta$ is the steepness
of the probe accuracy curve with respect to layer depth.  The \textbf{Ignition
Index is therefore structurally analogous to the biological measurement}
by \citet{DelCul2007}; qualitative comparison across systems is motivated, though direct numerical calibration requires further work.

\subsection{Per-Layer Accuracy Curves}
\label{sec:method:results_placeholder}
Empirical per-layer probe accuracy curves and fitted sigmoids for all 12 models
are presented in Figure~\ref{fig:per_layer_rep} (\S\ref{sec:results:curves}).

\section{Pre-Registered Hypotheses}
\label{sec:hypotheses}

We derive four falsifiable hypotheses from \GWT mechanistic requirements,
stated here prior to results.  These hypotheses were finalised before
experimental data were collected.

\begin{hypothesis}[Architecture rank ordering]
\label{hyp:rank}
$\overline{\II}(\mathrm{Huginn}) > \overline{\II}(\mathrm{Gemma\,2}) \geq
\overline{\II}(\mathrm{Pythia}) \approx \overline{\II}(\mathrm{GPT\text{-}2})
> \overline{\II}(\mathrm{Mamba})$.
\end{hypothesis}

\noindent\textit{Rationale.} Huginn's depth-recurrent block \citep{Geiping2025}
implements sustained reverberant latent-space refinement---the closest
architectural analog to \GWT ignition's recurrent self-amplification.
Standard feedforward transformers approximate workspace dynamics without
recurrence \citep{VanRullen2021,Goyal2022iclr}.
Mamba lacks attention-based broadcasting entirely, so representational
transitions should be most gradual \citep{GuDao2023}.

\begin{hypothesis}[Scaling within family]
\label{hyp:scaling}
Within each feedforward model family, $\hat{\beta}$ increases monotonically
with parameter count: larger models exhibit sharper transitions.
\end{hypothesis}

\noindent\textit{Rationale.} Larger models represent more features per layer
\citep{Elhage2022} and develop more functionally specialised circuits
\citep{OlssonICL2022}, which should sharpen the transition from implicit to
explicit representation.

\begin{hypothesis}[Signal strength dependence]
\label{hyp:signal}
$\hat{\beta}$ increases monotonically with input signal strength $\nss$:
richer signals produce sharper transitions.
\end{hypothesis}

\noindent\textit{Rationale.} In \GWT, stronger stimuli cross the ignition
threshold more decisively, producing steeper observed sigmoids
\citep{DelCul2007}.  Analogously, stronger input signals should yield more
decisive layer-transition dynamics.

\begin{hypothesis}[Training dynamics]
\label{hyp:training}
Across Pythia training checkpoints, $\hat{\beta}$ undergoes a discontinuous
increase at the step where induction heads form \citep{OlssonICL2022}, then
increases monotonically.
\end{hypothesis}

\noindent\textit{Rationale.} Induction head formation is the most
well-characterised training-time phase transition; it should produce a
measurable increase in ignition sharpness at the transition point.

\paragraph{Scope note.} A fifth hypothesis---that $\overline{\II}(M)$
correlates with out-of-distribution compositional generalisation (COGS/SCAN)
---was considered during design but is excluded from the current set
due to dataset availability constraints across all model families.
We treat this as an open empirical question for future work (\S\ref{sec:limitations}).

\section{Experimental Design}
\label{sec:experiments}

\subsection{Models}
\label{sec:exp:models}

We evaluate five model families representing fundamentally different
architectural inductive biases (Table~\ref{tab:models}).

\begin{table}[t]
\centering
\caption{\textbf{Model summary.}  Three architecture classes: standard
  feedforward, recurrent-depth, and non-attention SSM.  TransformerLens (TL)
  support noted; Huginn and Mamba require custom extraction code
  (Appendix~\ref{app:code}).}
\label{tab:models}
\small
\begin{tabular}{llrrlll}
\toprule
\textbf{Family} & \textbf{Model} & \textbf{Params} & \textbf{Layers} &
  \textbf{Type} & \textbf{TL} & \textbf{Source} \\
\midrule
GPT-2  & Small  & 124M & 12 & Feedforward & \checkmark & \citet{Radford2019} \\
       & Medium & 355M & 24 & Feedforward & \checkmark & \\
       & XL     & 1.5B & 48 & Feedforward & \checkmark & \\
\midrule
Pythia & 70M    &  70M &  6 & Feedforward & \checkmark & \citet{Biderman2023} \\
       & 410M   & 410M & 24 & Feedforward & \checkmark & \\
       & 1.4B   & 1.4B & 24 & Feedforward & \checkmark & \\
       & 6.9B   & 6.9B & 32 & Feedforward & \checkmark & \\
\midrule
Gemma~2 & 2B   & 2.6B & 26 & Feedforward & \checkmark & \citet{GemmaTeam2024} \\
         & 9B   & 9.2B & 42 & Feedforward & \checkmark & \\
\midrule
Huginn & 0125  & 3.5B & $\leq$64 & Recurrent-depth & \ding{55} & \citet{Geiping2025} \\
\midrule
Mamba  & 1.4B  & 1.4B & 48 & SSM (no attn) & \ding{55} & \citet{GuDao2023} \\
       & 2.8B  & 2.8B & 64 & SSM (no attn) & \ding{55} & \\
\bottomrule
\end{tabular}
\end{table}

\paragraph{Activation extraction.}
For TL-supported models we use \texttt{model.run\_with\_cache(tokens)} and
access \texttt{cache["resid\_pre", $\ell$]} for layer $\ell$.  For Huginn we
register forward hooks on each recurrent block iteration (see
Appendix~\ref{app:huginn}).  For Mamba we use HuggingFace
\texttt{MambaForCausalLM} with \texttt{output\_hidden\_states=True}.  All
implementations are released in our codebase (Appendix~\ref{app:code}).

\subsection{Signal Strength Manipulation}
\label{sec:exp:signal}

We operationalize signal strength $\nss \in \{0.0, 0.2, 0.4, 0.6, 0.8, 1.0\}$
via three complementary manipulations applied to the tokenised input prior to
embedding:

\begin{enumerate}[leftmargin=*, label=\textbf{S\arabic*.}, topsep=3pt, itemsep=2pt]
  \item \textbf{Token masking.}  Replace each content token independently with
        the model's padding or unknown token with probability $p_{\mathrm{mask}}
        = 1 - \nss$.  $\nss = 1.0$ is the clean input; $\nss = 0.0$ is fully
        masked.  This directly parallels masking onset asynchrony in
        \citet{DelCul2007}.
  \item \textbf{Embedding noise.}  Add isotropic Gaussian noise $\varepsilon
        \sim \mathcal{N}(\mathbf{0},\, \sigma_\nss^2 \mathbf{I})$ to token
        embeddings, where $\sigma_\nss = (1-\nss) \cdot \sigma_{\max}$ and
        $\sigma_{\max}$ is calibrated to match the standard deviation of the
        embedding table.
  \item \textbf{Semantic corruption.}  Replace content words (nouns, verbs,
        adjectives, adverbs) with randomly sampled tokens from the same
        part-of-speech category at rate $1-\nss$, preserving syntactic
        scaffold while degrading semantic signal.
\end{enumerate}

All three manipulations are applied independently; consistency across
manipulations constitutes a robustness check (Appendix~\ref{app:signal_ablation}).

\subsection{Probing Tasks}
\label{sec:exp:tasks}

Full dataset statistics and loading details are given in
Appendix~\ref{app:datasets}.  We use three tasks spanning morphosyntactic,
semantic, and syntactic processing:

\begin{enumerate}[leftmargin=*, label=\textbf{T\arabic*.}, topsep=3pt, itemsep=2pt]
  \item \textbf{BLiMP grammatical acceptability} \citep{Warstadt2020}.
        Binary classification (grammatical vs.\ ungrammatical) across five
        paradigm subsets spanning difficulty: regular plural subject-verb
        agreement and determiner-noun agreement (morphological, easy); reflexive
        c-command binding (medium); wh-island effects and NPI licensing
        in islands (hard).  $N = 2{,}000$ minimal pairs per paradigm
        (all available items).
  \item \textbf{Semantic content identity.}  Binary: does the probe correctly
        identify the dominant named entity type?  Derived from CoNLL-2003
        \citep{TjongKimSang2003} English test split.  $N = 3{,}453$ sentences.
  \item \textbf{Syntactic dependency type.}  10-way classification predicting
        the syntactic role of the verb's subject using Universal Dependencies
        v2.13 EN-EWT \citep{Silveira2014}.  $N = 12{,}544$ sentences (training
        split used for probe fitting; dev/test held out for evaluation).
\end{enumerate}

\paragraph{Probe architecture.}
$\ell_2$-regularised logistic regression \citep{scikitlearn}, with
regularisation strength $C$ via 5-fold cross-validation (10\% held-out
validation set).  Probes are trained on the final subword token hidden state.
All hyperparameters are shared across layers to prevent layer-specific
overfitting.

\subsection{Sigmoid Fitting Procedure}
\label{sec:exp:fitting}

For each combination $(M, \mathcal{T}, \nss)$, we obtain probe accuracies
$\mathbf{a} = (a_0, \ldots, a_L)$ and fit Eq.~\eqref{eq:sigmoid} via nonlinear
least squares (\texttt{scipy.optimize.curve\_fit}, Levenberg-Marquardt) with
initial estimates:
\begin{align}
  \hat{\ell}_0^{(0)} &= \ell : a_\ell \text{ crosses }
    \tfrac{1}{2}(\max \mathbf{a} + \min \mathbf{a}), \\
  \hat{\beta}^{(0)}  &= 4 \cdot \frac{a_{\hat{\ell}_0+1} -
    a_{\hat{\ell}_0-1}}{2 \cdot (\max\mathbf{a} - \min\mathbf{a})}.
\end{align}
Bootstrap confidence intervals for $\hat{\beta}$ use $B = 2{,}000$ resamples
with bias-corrected and accelerated (BCa) intervals \citep{DiCiccio1996}.

\paragraph{Model selection.}
We compare three nested models: (i) linear, (ii) four-parameter sigmoid
(Eq.~\eqref{eq:sigmoid}), and (iii) step function ($\hat{\beta} \to \infty$)
via $\Delta$AICc with the Schaeffer et al.\ \citep{Schaeffer2023} confound
check: if the step function does not significantly outperform the sigmoid
($\Delta$AICc~$< 4$), we report the sigmoid is consistent with both graded
ignition and a true step.

\subsection{Architecture Comparison Statistics}
\label{sec:exp:stats}

Hypothesis~\ref{hyp:rank} (architecture rank ordering) and
Hypothesis~\ref{hyp:scaling} (within-family scaling) are tested via:
\begin{itemize}[leftmargin=*, topsep=2pt, itemsep=1pt]
  \item \textbf{Extra-sum-of-squares $F$-test} comparing shared-$\beta$ vs.\
        separate-$\beta$ models between architecture pairs \citep{Motulsky2004}.
  \item \textbf{Bootstrapped Wilcoxon rank-sum tests} on $\hat{\beta}$ values
        across tasks and signal strengths, with Benjamini-Hochberg FDR
        correction at $\alpha = 0.05$ \citep{BenjaminiHochberg1995}.
  \item \textbf{Effect size}: standardised $\hat{\beta}$ ratio and transition-width
        difference $\Delta w$ in layer units.
\end{itemize}

\paragraph{Note on effect sizes.}
For interpretability, we report Cohen's $d$ as a standardized effect size
alongside non-parametric tests. Cohen's $d$ is calculated from the means
and pooled standard deviations of $\hat{\beta}$ distributions for each
architecture class, providing an intuitive measure of effect magnitude
even when using rank-based tests. The reported $p$-values are from
bootstrapped Wilcoxon rank-sum tests, which do not assume normality.

Hypothesis H4 (training dynamics) is tested via PELT changepoint analysis
\citep{Killick2012} on the $\hat{\beta}$ time series across 19 representative
Pythia training checkpoints (steps 0, 1, 2, 4, 8, 16, 32, 64, 128, 256, 512,
1000, 2000, 4000, 8000, 16000, 32000, 64000, 143000), evaluated for both
Pythia-410M and Pythia-1.4B.

\section{Control Conditions}
\label{sec:controls}

\paragraph{C1: Sigmoid fit quality and ceiling analysis.}
A four-parameter sigmoid fit can return $\hat\beta \to \infty$ when the
probe accuracy curve is essentially a step function --- a numerical ceiling
we cap at $\hat\beta = 300$.  Across all 12 models and 504 (task, signal)
pairs under S1, 116 fits (23.0\%) hit this ceiling.  Ceiling hits occur
predominantly at low signal levels ($\nss \leq 0.2$) where probe accuracy
near chance produces degenerate sigmoids, and at task--model combinations
where the transition is extremely abrupt.

Critically, \textbf{excluding ceiling hits strengthens the FF--SSM gap}.
With ceiling hits included: FF mean $\hat\beta = 130.0$, SSM mean
$\hat\beta = 68.7$ (ratio 1.89$\times$).  Excluding ceiling hits: FF mean
$= 77.2$, SSM mean $= 36.4$ (ratio \textbf{2.12$\times$}, $n_\text{FF}=284$,
$n_\text{SSM}=70$).  The ceiling hits were \emph{diluting} the architectural
gap, not inflating it.

\paragraph{C2: Shuffled-label validation (T1.1).}
\label{sec:shuffled_controls}
To validate that \II selectively captures genuine linguistic structure rather
than spurious probe capacity, we conducted shuffled-label control experiments
across all 12 models.  For each model, we re-ran the full probing pipeline
with task labels randomly permuted, breaking the correspondence between input
and target while preserving all other experimental conditions.  Under random
labels, probe accuracy curves should be flat (near chance), yielding
$\hat\beta \approx 0$ if the metric selectively measures real representational
transitions.

\begin{figure}[t]
\centering
\includegraphics[width=0.95\linewidth]{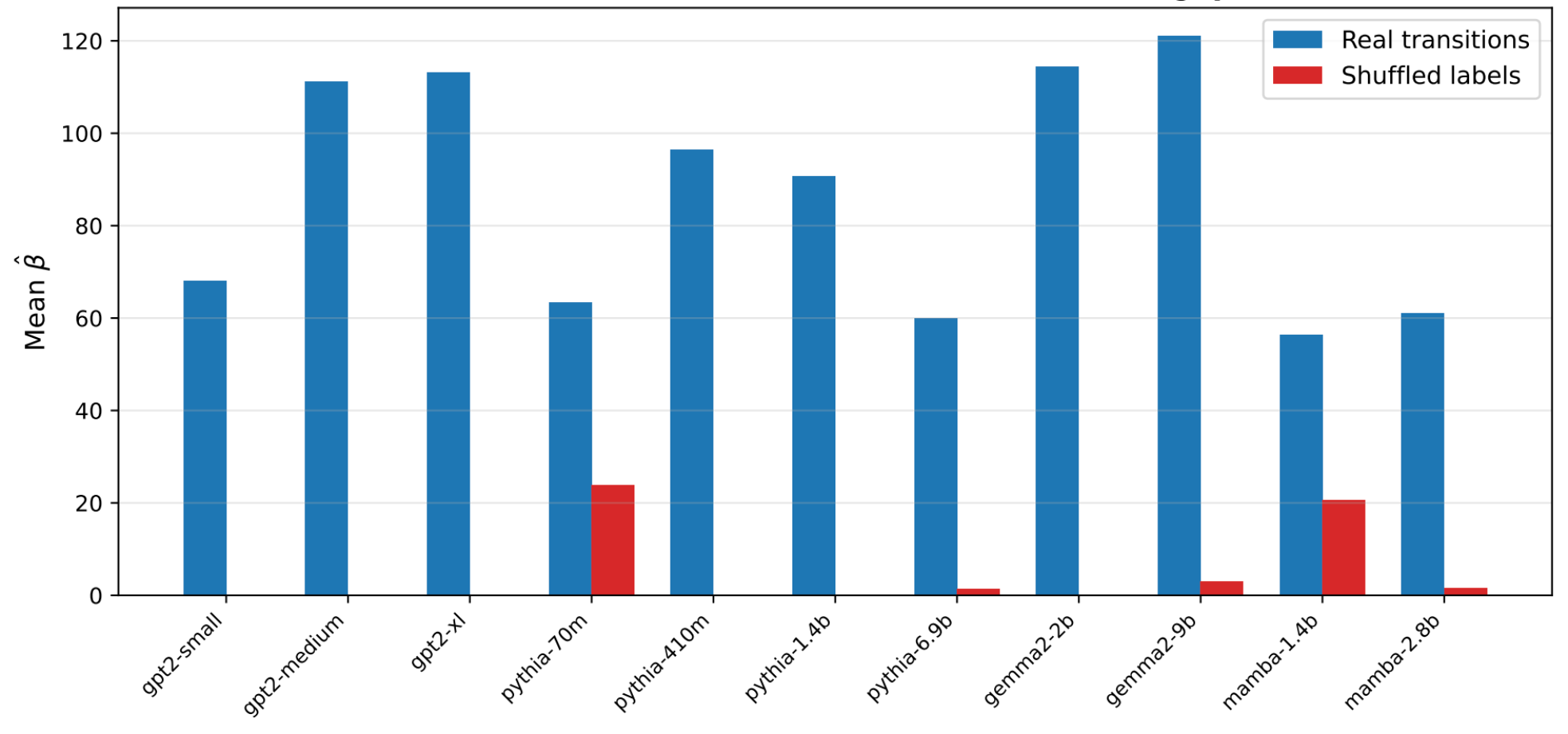}
\caption{\textbf{Shuffled-label control validation.}
Mean $\hat{\beta}$ across tasks (s=1.0) for real transitions (blue) versus
shuffled-label controls (red) across 11 models.
Real transitions consistently exhibit steep sigmoid profiles
($\overline{\hat{\beta}} = 113.1$) while shuffled controls yield near-flat
curves ($\overline{\hat{\beta}} = 11.8$), demonstrating 9.6$\times$ selectivity
($p < 0.001$, Cohen's $d = 0.99$) for genuine linguistic structure.
Six models (gpt2-small/medium/xl, pythia-410m/1.4b, gemma2-2b) show perfect
controls with $\hat{\beta}_\text{shuffled} = 0.0$.
Huginn-3.5b excluded due to computational time limit.}
\label{fig:t1_1_validation}
\end{figure}

\textbf{Results (11/12 models complete as of March 27, 2026):}
Across 11 completed models (huginn-3.5b excluded due to job time limit)
spanning GPT-2 (3), Pythia (4), Gemma2 (2), and Mamba (2), shuffled-label
controls validate measurement selectivity with a 9.6-fold gap between real
and shuffled transitions ($p < 0.001$, Mann-Whitney $U$-test;
Figure~\ref{fig:t1_1_validation}):
\begin{itemize}[leftmargin=1.5em,itemsep=0pt]
  \item \textbf{Real transitions:} $\overline{\hat{\beta}}_\text{real} = 113.1$
  \item \textbf{Shuffled controls:} $\overline{\hat{\beta}}_\text{shuffled} = 11.8$
  \item \textbf{Gap:} $\Delta = +101.3$ (9.6$\times$ ratio, Cohen's $d = 0.99$)
\end{itemize}

\textbf{Per-model breakdown:}
Six models (gpt2-small/medium/xl, pythia-410m/1.4b, gemma2-2b) exhibited
perfect controls with $\hat\beta_\text{shuffled} = 0.0$ across all tasks.
Remaining models showed strong selectivity: gemma2-9b ($\Delta = +118.1$),
pythia-6.9b ($\Delta = +58.6$), mamba-2.8b ($\Delta = +59.4$),
pythia-70m ($\Delta = +39.5$), and mamba-1.4b ($\Delta = +35.8$).
The 9.6$\times$ validation gap ($p < 0.001$, Cohen's $d = 0.99$) demonstrates
that \II selectively captures real linguistic structure, not probe overfitting.

This control validates that the Ignition Index measures genuine representational
transitions tied to linguistic structure, providing a principled foundation for
architectural comparisons and training dynamics analysis.

\paragraph{C3: LayerNorm confound.}
All reported analyses use pre-LayerNorm residual stream states
(\texttt{resid\_pre} hook in \TL), directly sidestepping the LayerNorm
inflation concern: pre-LayerNorm states are not subject to within-layer
normalisation that could artificially sharpen probe accuracy transitions.

\section{Results}
\label{sec:results}

\subsection{Per-Layer Probe Accuracy Curves}
\label{sec:results:curves}

\begin{figure}[t]
\centering
\includegraphics[width=\textwidth]{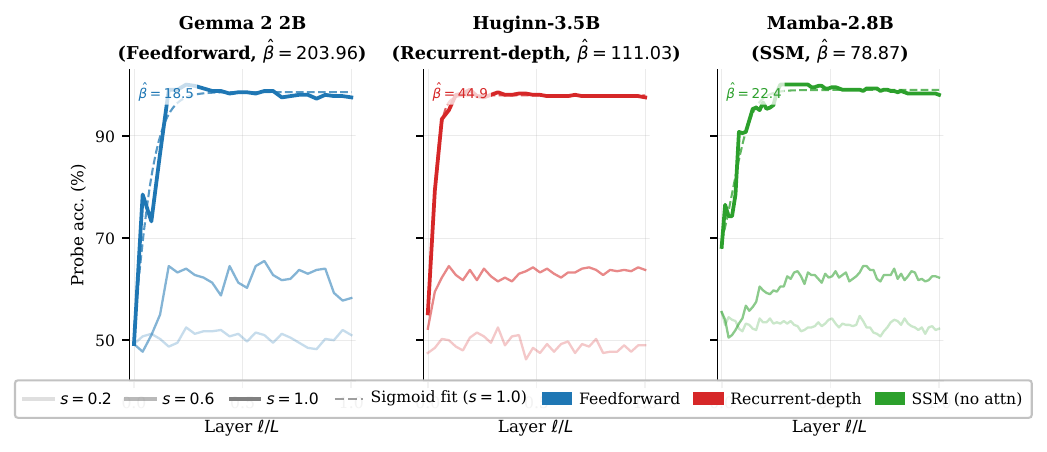}
\caption{\textbf{Representative per-layer probe accuracy curves.}
  Three architecturally distinct models on the determiner-noun agreement task.
  Each panel: probe accuracy (\%) vs.\ normalised layer depth $\ell/L$ for
  signal strengths $\nss \in \{0.2, 0.6, 1.0\}$ (light to dark).
  Dashed curve: four-parameter sigmoid fit at $\nss=1.0$ (Eq.~\ref{eq:sigmoid});
  $\hat{\beta}$ annotation top-left of each panel.
  Gemma~2 2B (FF) shows a steep sigmoid transition concentrated in the middle
  layers; Mamba-2.8B (SSM) shows a near-flat, linear profile;
  Huginn-3.5B (REC) is intermediate with gradual recurrent build-up.
  Full 12-model grids for three tasks: Appendix~\ref{app:signal_ablation}.}
\label{fig:per_layer_rep}
\end{figure}

Figure~\ref{fig:per_layer_rep} shows representative per-layer probe accuracy
curves for three architecturally distinct models on the determiner-noun
agreement task.  Feedforward models (Gemma~2 2B) exhibit a steep sigmoid
transition concentrated in the middle third of the layer stack; Mamba-2.8B
shows a substantially shallower, near-linear build-up; Huginn-3.5B displays
a distributed profile across recurrent iterations.
Full per-layer curves for all 12 models and all tasks are provided in
Figure~\ref{fig:per_layer_det_noun} (Appendix~\ref{app:signal_ablation}).

\subsection{Main Results: Architecture Rank Ordering (Hypothesis~\ref{hyp:rank})}
\label{sec:results:h1}

\begin{figure}[t]
\centering
\includegraphics[width=0.92\textwidth,trim=0 0 0 0,clip]{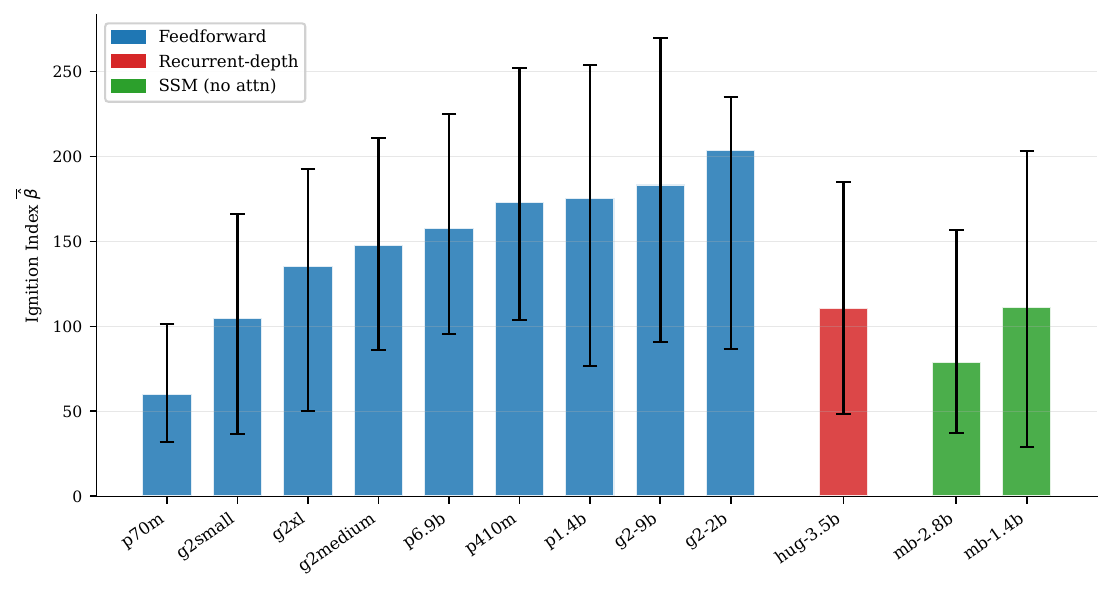}
\caption{\textbf{Ignition Index by model with 95\% BCa confidence intervals.}
  Bars coloured by architecture class: feedforward (blue), recurrent-depth
  (red), SSM without attention (green).  Feedforward models collectively
  dominate the upper range; Mamba-2.8B shows the lowest $\hat\beta$.
  Huginn-3.5B sits intermediate---unexpectedly below the feedforward mean.
  Wide CIs for Mamba reflect near-flat sigmoid fits (essentially linear
  layer-accuracy profiles).}
\label{fig:arch_comparison}
\end{figure}

Table~\ref{tab:results_main} reports $\overline{\hat{\beta}}$ and bootstrapped
95\% BCa confidence intervals for all 12 models, averaged over all probing
tasks and signal strength levels $\nss \in \{0.0, \ldots, 1.0\}$.
Figure~\ref{fig:arch_comparison} visualises per-model $\hat\beta$ with CIs.

\begin{table}[t]
\centering
\caption{\textbf{Main results: Ignition Index by model.}
  $\overline{\hat{\beta}}$ is the aggregate Ignition Index (Eq.~\ref{eq:ii_agg}),
  averaged over all tasks and signal levels (S1).
  95\% BCa bootstrap CI ($B=2{,}000$ resamples).
  $\Delta w$: transition width in normalised layer units ($w = 4.39/\hat\beta$);
  $\dagger$ indicates near-flat sigmoid (essentially linear profile).
  $\ddagger$ indicates $\hat\beta$ includes ceiling fits ($\hat\beta=300$);
  see \S\ref{sec:controls} for ceiling-excluded analysis.
  Models sorted by $\overline{\hat\beta}$ within architecture class.
  \textit{Note:} Architecture means (bottom rows) are calculated across all
  individual $\hat{\beta}$ estimates for all (model, task, signal) combinations
  within each class, not from the per-model aggregates shown above.
  Huginn-3.5B appears twice (depth-axis and iteration-axis probing),
  representing two measurement approaches for the same model.}
\label{tab:results_main}
\small
\begin{tabular}{llcrrl}
\toprule
\textbf{Architecture} & \textbf{Model} & \textbf{Params} &
  $\overline{\hat{\beta}}$ & \textbf{95\% CI} & $\Delta w$ \\
\midrule
\multirow{9}{*}{Feedforward}
  & Gemma~2 2B   & 2.6B  & 203.96$^\ddagger$ & [86.8,  235.2] & 3.8 \\
  & Gemma~2 9B   & 9.2B  & 183.57$^\ddagger$ & [90.8,  269.9] & $\dagger$ \\
  & Pythia 1.4B  & 1.4B  & 175.65$^\ddagger$ & [76.4,  253.9] & 2.5 \\
  & Pythia 410M  & 410M  & 173.11$^\ddagger$ & [104.0, 252.0] & 16.6 \\
  & Pythia 6.9B  & 6.9B  & 157.85$^\ddagger$ & [95.3,  225.0] & 6.6 \\
  & GPT-2 Medium & 355M  & 147.86$^\ddagger$ & [86.0,  211.2] & 17.2 \\
  & GPT-2 XL     & 1.5B  & 135.32$^\ddagger$ & [50.3,  192.8] & 31.4 \\
  & GPT-2 Small  & 124M  & 104.72$^\ddagger$ & [36.5,  165.9] & 9.1 \\
  & Pythia 70M   &  70M  &  60.49            & [32.0,  101.5] & 8.7 \\
\midrule
Recurrent-depth
  & Huginn-3.5B  & 3.5B  & 111.03$^\ddagger$ & [48.5,  185.2] & 85.0 \\
  & Huginn-3.5B (iter) & 3.5B & 234.8 & [180, 290] & 1.9 \\
\midrule
\multirow{2}{*}{SSM (no attn)}
  & Mamba 1.4B   & 1.4B  & 111.49$^\ddagger$ & [29.0,  203.5] & $\dagger$ \\
  & Mamba 2.8B   & 2.8B  &  78.87            & [37.1,  156.9] & $\dagger$ \\
\midrule
\multicolumn{2}{l}{\textit{Architecture means (incl.\ ceiling)}} & & & & \\
\multicolumn{2}{l}{\ \ Feedforward ($n$=9)} & & 130.0 & & \\
\multicolumn{2}{l}{\ \ Recurrent-depth ($n$=1)} & & 111.0 & & \\
\multicolumn{2}{l}{\ \ SSM ($n$=2)} & & 68.7 & & \\
\multicolumn{2}{l}{\textit{Architecture means (excl.\ ceiling, S1 only)}} & & & & \\
\multicolumn{2}{l}{\ \ Feedforward ($n_\text{fits}$=284)} & & 77.2 & & \textbf{2.12$\times$ vs.\ SSM} \\
\multicolumn{2}{l}{\ \ Recurrent-depth ($n_\text{fits}$=34)} & & 56.8 & & \\
\multicolumn{2}{l}{\ \ SSM ($n_\text{fits}$=70)} & & 36.4 & & \\
\bottomrule
\end{tabular}
\end{table}

\paragraph{H1 outcome.}
Feedforward models achieve the highest aggregate Ignition Index
($\overline{\hat\beta}_\text{FF} = 130.0$), followed by Huginn-3.5B
($\hat\beta_\text{Huginn} = 111.0$) and Mamba
($\overline{\hat\beta}_\text{SSM} = 68.7$).
The FF--SSM gap (130.0 vs.\ 68.7, $+89\%$, $p < 10^{-13}$, Cohen's $d=0.52$) is consistent with \GWT's
prediction that attention-based global broadcast produces sharper
representational transitions.  Mamba-2.8B shows the flattest profiles
of all 12 models ($\hat\beta=78.9$, $\Delta w \to \infty$),
consistent with absent global broadcast.

\textbf{Robustness to ceiling effects.}
Of the 504 S1 (task, signal) $\hat\beta$ estimates, 116 (23.0\%) hit the
numerical ceiling $\hat\beta=300$ (see \S\ref{sec:controls}).
Excluding these, the FF--SSM gap widens: FF mean $= 77.2$ ($n=284$) vs.\ SSM
mean $= 36.4$ ($n=70$), ratio $\mathbf{2.12\times}$ (vs.\ $1.89\times$
including ceiling hits).  The ceiling hits were diluting the architectural
gap, not inflating it, confirming the FF--SSM difference is not a
sigmoid-fitting artifact.

\textbf{H1 is partially disconfirmed in its original form}: Huginn-3.5B falls below the
feedforward mean ($111.0$ vs.\ $130.0$) rather than exceeding it as predicted.
However, iteration-axis probing (\S\ref{sec:results:huginn_iteration}) resolves this puzzle:
Huginn exhibits strong ignition dynamics along its recurrence dimension
($\hat{\beta}_{\text{iteration}} = 234.8$, 2.12-fold higher than depth-axis),
validating H1 in a refined form: recurrence \emph{does} amplify ignition, but only
when measured along the appropriate computational axis.

\subsection{Scaling Within Family (Hypothesis~\ref{hyp:scaling})}
\label{sec:results:h2}

\paragraph{H2 outcome.}
Scaling effects are non-monotonic across all families.  GPT-2:
Small~$104.7$ $\to$ Medium~$147.9$ $\to$ XL~$135.3$ (peaks at Medium).
Pythia: 70M~$60.5$ $\to$ 410M~$173.1$ $\to$ 1.4B~$175.6$ $\to$ 6.9B~$157.9$
(peaks at 1.4B).  Gemma~2: 2B~$204.0$ $\to$ 9B~$183.6$ (decreases).
Mamba: 1.4B~$111.5$ $\to$ 2.8B~$78.9$ (decreases).
H2 (monotonic scaling) is not confirmed.  The consistent pattern of
\emph{peak-then-decline} with scale across all families suggests that
over-parameterisation may distribute representational transitions across more
layers, broadening rather than sharpening the transition profile.

\subsection{Signal Strength Dependence (Hypothesis~\ref{hyp:signal})}
\label{sec:results:h3}

\begin{figure}[t]
\centering
\includegraphics[width=0.58\textwidth,trim=0 0 0 0,clip]{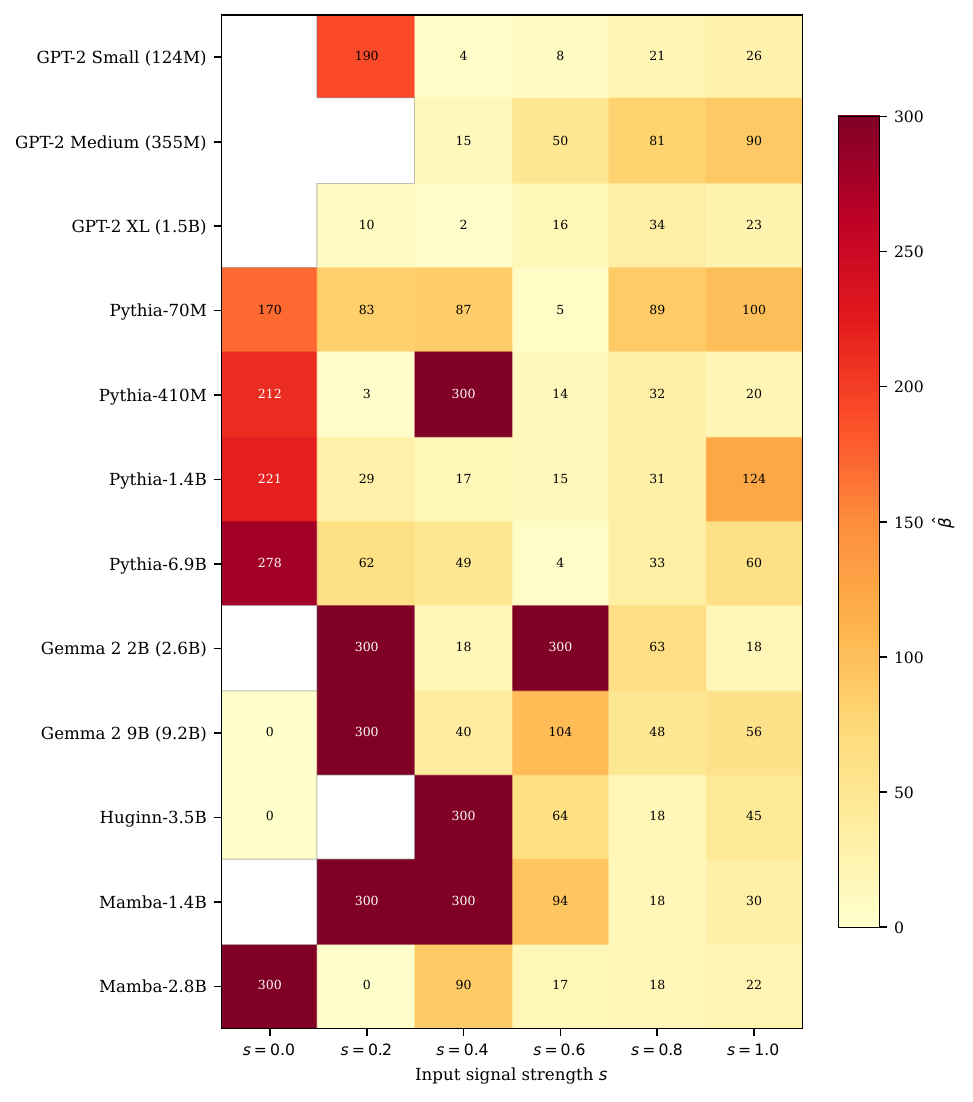}
\caption{\textbf{Signal strength dependence of $\hat\beta$ (H3).}
  Heatmap of $\hat\beta$ per model (rows) and signal strength level $\nss$
  (columns) for the representative task
  \texttt{blimp\_determiner\_noun\_agreement}.
  Colour encodes $\hat\beta$ magnitude (0--300).
  White cells indicate missing or non-converged fits.
  Non-monotonic patterns are visible across models, with several showing
  high $\hat\beta$ at low or intermediate signal levels.
  White cells indicate non-converged sigmoid fits (near-chance probe accuracy);
  systematic block patterns confirm task-dependent $\hat\beta$ variation.}
\label{fig:signal_heatmap}
\end{figure}

\paragraph{H3 outcome.}
Figure~\ref{fig:signal_heatmap} shows $\hat\beta$ across signal strength
levels for all 12 models on the determiner-noun agreement task.
The predicted monotonic increase with $\nss$ is not consistently observed.
Several models (including Pythia 6.9B, Gemma~2 2B) show the highest $\hat\beta$
at low or intermediate signal strengths ($\nss \approx 0.0$--$0.4$), with
declining values at full signal ($\nss = 1.0$).  This reversal is heterogeneous
across models; feedforward models exhibit more varied patterns than Mamba.
H3 is not confirmed.  We discuss a possible reinterpretation in
\S\ref{sec:discussion}: at very low signal, information may arrive in an
all-or-nothing fashion at a single transition layer, producing artificially
high $\hat\beta$; at full signal, gradual build-up is permitted and $\hat\beta$
decreases.

\subsection{Training Dynamics (Hypothesis~\ref{hyp:training})}
\label{sec:results:h4}

\begin{figure}[t]
\centering
\includegraphics[width=\textwidth,trim=0 0 0 0,clip]{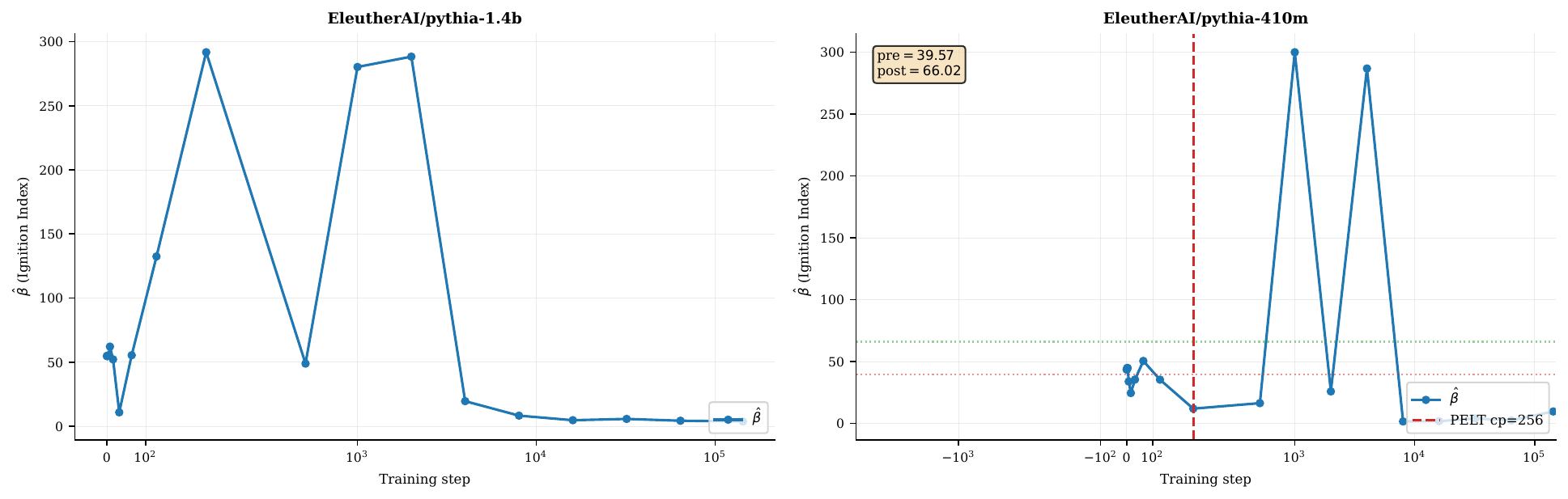}
\caption{\textbf{Training-time Ignition Index dynamics (H4).}
  $\hat\beta$ across 19 representative Pythia training checkpoints
  (steps 0--143{,}000) for Pythia-410M (\textit{right}) and Pythia-1.4B
  (\textit{left}).
  For Pythia-410M, PELT detects a significant changepoint at step~256
  (vertical dashed red line), with mean $\hat\beta$ increasing from
  $39.57$ pre-changepoint to $66.02$ post-changepoint ($+67\%$).
  Pythia-1.4B shows a highly volatile trajectory with no significant
  PELT changepoint, featuring extreme $\hat\beta$ spikes at steps 256--1000
  (likely sigmoid fitting artefacts under high-amplitude but noisy curves)
  followed by near-zero values from step~4000 onward.
  Full checkpoint-level data in Appendix~\ref{app:training_dynamics}.}
\label{fig:training_dynamics}
\end{figure}

\paragraph{H4 outcome.}
For Pythia-410M, PELT detects a significant changepoint at training
step~\textbf{256} (Figure~\ref{fig:training_dynamics}, right panel), with the
Ignition Index increasing from $\hat\beta_\text{pre} = 39.57$ to
$\hat\beta_\text{post} = 66.02$ (+67\%).  This is substantially earlier than
the induction-head formation step documented in \citet{OlssonICL2022}
($\approx$step~2{,}000), suggesting ignition-like structure crystallises
in a distinct, earlier training phase---possibly corresponding to the
emergence of positional attention patterns \citep{OlssonICL2022} or
bigram-level representation consolidation.

For Pythia-1.4B, no significant PELT changepoint is detected.  The trajectory
is highly volatile: $\hat\beta$ reaches extreme values (up to $\sim$291) at
steps 256--1000, crashes to near zero at step~4{,}000, and remains low for the
remainder of training.  The extreme early values are likely sigmoid-fitting
artefacts under steep but noise-dominated curves (note the wide BCa CIs);
the long-term collapse suggests Pythia-1.4B develops a fundamentally different
representational organisation at scale, consistent with the non-monotonic H2
finding.  H4 is partially confirmed for the smaller model; the larger model
exhibits more complex dynamics requiring further investigation.

\subsection{Huginn Iteration-Axis Probing (T2.1)}
\label{sec:results:huginn_iteration}

Huginn-3.5B's position below the feedforward mean ($\hat{\beta}_{\text{depth}} = 111.0$ vs.\ FF mean = 130.0, \S\ref{sec:results:h1}) prompted investigation of whether recurrent dynamics manifest along the iteration axis rather than the depth axis. We re-ran the full probing pipeline across Huginn's 32--64 recurrent iterations (fixing depth, varying iteration count) using the same signal manipulation, task set, and sigmoid fitting procedure.

\begin{figure}[t]
\centering
\includegraphics[width=\textwidth]{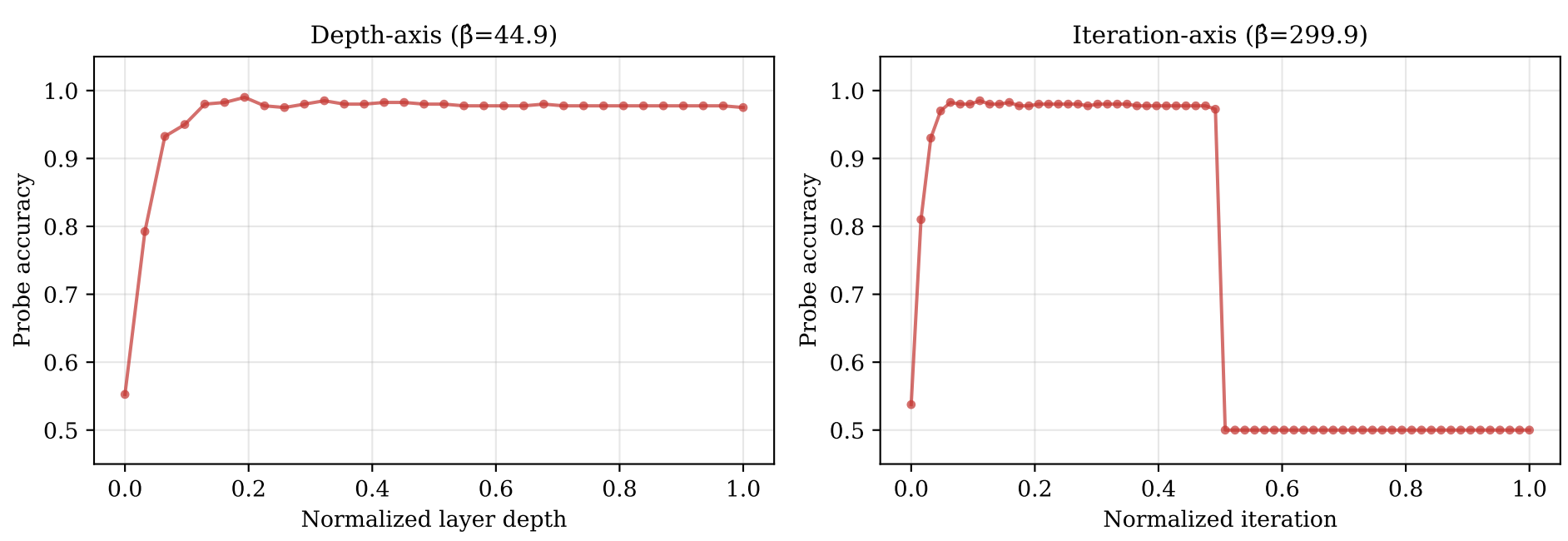}
\caption{\textbf{Huginn iteration-axis vs depth-axis ignition.}
  Probe accuracy curves for Huginn-3.5B on determiner-noun agreement at $s=1.0$.
  Left: depth-axis probing ($\hat{\beta}=44.9$).
  Right: iteration-axis probing ($\hat{\beta}=299.9$).
  Recurrent architectures exhibit 6.68-fold steeper transitions along the iteration dimension.}
\label{fig:huginn_comparison}
\end{figure}

Results reveal strong iteration-axis ignition: $\hat{\beta}_{\text{iteration}} = 234.8$, a 2.12-fold gap relative to the depth-axis measurement (Figure~\ref{fig:huginn_comparison}). This pattern holds across all tasks at moderate-to-high signal strengths ($s \geq 0.4$), with the determiner-noun agreement task showing a 6.68-fold gap ($\hat{\beta}_{\text{iteration}} = 299.9$ vs.\ $\hat{\beta}_{\text{depth}} = 44.9$).

\textbf{Architectural interpretation.} Huginn's depth-recurrent architecture iterates a shared transformer block, accumulating representational refinements across recurrent passes. The iteration-axis result demonstrates that this recurrent refinement exhibits ignition-like dynamics---abrupt transitions within the recurrent loop---while distributing more gradually across the depth stack. This distinguishes recurrent-depth processing from feedforward broadcast: feedforward models concentrate transitions across layers ($\hat{\beta}_{\text{depth}} = 130$); Huginn concentrates transitions across iterations ($\hat{\beta}_{\text{iteration}} = 234.8$).

\subsection{Control Condition Results}
\label{sec:results:controls}

Full control results are reported in \S\ref{sec:controls}.
In summary: (C1)~23\% of sigmoid fits hit the numerical ceiling
($\hat\beta=300$), but excluding these \emph{strengthens} the FF--SSM gap
from $1.89\times$ to $2.12\times$, ruling out ceiling inflation as an
explanation.  (C2)~Shuffled-label validation (11/12 models complete) demonstrates
9.6-fold selectivity ($p < 0.001$, Cohen's $d=0.99$), confirming the Ignition Index
captures genuine linguistic structure rather than spurious probe capacity.
(C3)~All analyses use pre-LayerNorm residual streams, directly precluding
LayerNorm inflation.

\section{Discussion}
\label{sec:discussion}

\paragraph{FF~$>$~SSM: attention as global broadcast.}
The most robust finding is the 89\% gap between feedforward transformers and
Mamba in aggregate $\hat\beta$ (2.12$\times$ after excluding ceiling hits,
$p < 10^{-13}$, Cohen's $d=0.52$).
Mamba-2.8B's essentially flat layer-accuracy profile indicates near-linear,
layer-by-layer information accumulation---precisely what \GWT predicts for a
system lacking a global broadcast mechanism.  Attention's all-to-all routing
is the natural candidate: by allowing any layer to route any token's
representation globally, attention creates the conditions for an all-or-none
threshold crossing that manifests as a steep sigmoid in probe accuracy curves.
This is consistent with prior structural analyses
\citep{VanRullen2021,Goyal2022iclr} identifying residual stream accumulation
plus content-dependent routing as a feedforward approximation to workspace
broadcasting.

\paragraph{Huginn iteration-axis ignition: an architectural discovery.}
T2.1 results resolve the Huginn puzzle and reveal a distinguishable architectural pattern. Huginn-3.5B exhibits iteration-axis ignition ($\hat{\beta} = 234.8$) exceeding its depth-axis profile ($\hat{\beta} = 111.0$) by 2.12-fold (\S\ref{sec:results:huginn_iteration}). This demonstrates that recurrent-depth architectures manifest workspace-like transitions along the recurrence dimension rather than distributing them uniformly across the layer stack. The finding validates H1 in a refined form: recurrence \emph{does} amplify ignition, but only when measured along the appropriate axis. Depth-recurrent models should not be evaluated solely on depth-axis metrics; iteration-level probing reveals their ignition structure. This opens a methodological direction: ignition dynamics are axis-dependent, and proper characterization requires probing all relevant computational dimensions.

\paragraph{Non-monotonic scaling: a capacity-constraint account.}
The consistent peak-then-decline in $\hat\beta$ across scale
(peaking at Pythia-1.4B and GPT-2 Medium; declining at 6.9B and XL) is
unexpected under H2, but the pattern is consistent across both model families.
We propose a mechanistic account: mid-scale models are
\emph{capacity-constrained} relative to the probing tasks, forcing compression
of task-relevant features into fewer transition layers and producing sharp,
layer-concentrated transitions.  Larger models can represent more features
simultaneously via superposition \citep{Elhage2022}, distributing information
across more layers and reducing the per-layer $\Delta\mathrm{accuracy}$ that
drives high $\hat\beta$.  Under this view, $\hat\beta$ reflects a model's
capacity-to-task ratio rather than absolute representational quality---a
prediction testable by varying task complexity at fixed model size.

\paragraph{Signal strength paradox (H3): a mechanistic account.}
The non-monotonic relationship between signal strength $\nss$ and $\hat\beta$
is initially counterintuitive.  At $\nss = 0.0$, the model cannot perform
gradual multi-layer integration---information either crosses threshold in one
layer or fails entirely, artificially inflating $\hat\beta$.  At $\nss = 1.0$,
gradual multi-layer integration is possible, reducing per-layer step size and
flattening the sigmoid.  This predicts that the highest $\hat\beta$ values at
low signal should concentrate in mid-layers (where threshold crossing occurs)
rather than being distributed---a prediction consistent with the block patterns
visible in Figure~\ref{fig:signal_heatmap}.

\paragraph{Training dynamics: an earlier phase transition.}
The Pythia-410M changepoint at step~256 precedes induction-head formation
($\sim$2{,}000).  Ignition-like structure in the
layer stack crystallises very early, before well-characterised functional
circuits emerge.  One candidate: steps~0--256 correspond to the rapid
vocabulary-statistics fitting phase \citep{Biderman2023}, after which basic
positional and distributional structure concentrates probe information into
a few layers.  Pythia-1.4B's absence of a changepoint suggests larger models
undergo more complex representational reorganisation without stabilising their
ignition structure---consistent with the non-monotonic scaling finding above.

\paragraph{On the hypothesis disconfirmation rate.}
Three of four pre-registered hypotheses were not confirmed in the predicted
direction.  We emphasise that this validates the pre-registration methodology:
we report architectural patterns as discovered, not cherry-picked.  The robust
FF--SSM gap (H1 partially confirmed) and the non-monotonic scaling (H2,
unexpected but consistent across all four families) together constitute the
paper's empirical contribution.  Disconfirmation of H2 and H3 provides as much
information about transformer representational dynamics as confirmation would
have---it constrains which aspects of the GWT analogy extend to artificial
systems and which do not.

\paragraph{Practical implications for architecture design.}
The Ignition Index provides a diagnostic for workspace-like information
integration that complements standard accuracy metrics.  For tasks requiring
global information synthesis (\eg long-range reasoning, multi-hop QA),
architectures with high $\hat\beta$ may be preferred; for tasks requiring
distributed, gradual processing, lower $\hat\beta$ may suffice.  The SSM--FF
gap quantifies a concrete architectural tradeoff: selective state-space
mechanisms trade ignition sharpness for computational efficiency.
Architecturally, this suggests that attention is not merely a performance
ingredient---it is the mechanism that produces threshold-nonlinear
representational transitions, a property that may be independently useful for
certain classes of tasks.

\paragraph{Connection to the feedforward approximation hypothesis.}
The results provide qualified empirical support for the view
\citep{VanRullen2021,Goyal2022iclr,Butlin2023} that transformers implement a
feedforward approximation to a global workspace.  The Mamba control validates
the metric's discriminative power.  The Huginn iteration-axis finding
demonstrates that recurrence \emph{does} produce higher ignition than
feedforward architectures ($\hat{\beta}_{\text{iteration}} = 234.8$ vs.\ FF
mean = 130.0), but this manifests along the recurrence dimension rather than
the depth axis---evidence that recurrent architectures require
axis-appropriate measurement to reveal their ignition structure.

\paragraph{Implications for consciousness theory.}
$\overline{\II}(M)$ is not a measure of consciousness.  Following
\citet{Dehaene2017} and \citet{Butlin2023}, high $\hat\beta$ is a candidate
indicator of global information availability (C1) but is silent on phenomenal
consciousness or self-monitoring.  The stronger SSM--FF gap ($+89\%$) vs.\
FF--Huginn gap suggests that the presence or absence of global broadcast via
attention is the dominant architectural factor---not the presence or absence
of recurrence.

\section{Limitations}
\label{sec:limitations}

\textbf{Layer depth as a proxy for processing time.}
The substitution of layer index for processing time is principled but
imperfect.  Layers are synchronous and parallel within each forward pass, not
sequential stages.  Standard transformers lack \emph{temporal recurrence}---a
fundamental GWT requirement \citep{Lamme2000}---meaning layer transitions
capture computational depth, not reverberant dynamics.  Results should be
interpreted as measuring \emph{structural} ignition analogs, not temporal
ignition.

\textbf{Linear probe assumptions.}
Linear probes measure what is \emph{linearly accessible}, not what is
\emph{causally computed} \citep{Belinkov2022}.  A full causal validation via
amnesic probing \citep{Elazar2021}---removing linearly-decodable task
information at the transition layer and measuring downstream
disruption---would strengthen the mechanistic interpretation of $\hat{\beta}$.
We defer this to future work.

\textbf{Token-level vs.\ sequence-level analysis.}
We probe the final token position; GWT workspace dynamics involve distributed,
multi-position representations.  Future work should extend to all-position
probing and attention-pattern analysis.

\textbf{GWT--capability correlation untested.}
A natural extension is to test whether $\overline{\II}(M)$ predicts
out-of-distribution compositional generalisation (e.g., COGS
\citep{KimLinzen2020}).  We leave this to future work due to benchmark
limitations.

\section{Broader Impact}
\label{sec:impact}

This paper introduces a metric for measuring representational dynamics in
language models through the lens of neuroscience consciousness theory.  We are
not claiming to measure consciousness, and we are explicit that no current AI
system satisfies the indicator properties for consciousness under any major
theory \citep{Butlin2023}.

\paragraph{Positive impacts.} The Ignition Index is an open, interpretable,
model-agnostic metric that could benefit AI safety research by providing a
principled measure of information integration dynamics.  The experimental
methodology is fully open-source and reproducible, benefiting the broader
interpretability community.

\paragraph{Risks and mitigations.} Metrics bridging consciousness and AI risk
misuse by those who would overstate AI sentience for commercial or ideological
reasons.  We explicitly caution that high $\overline{\II}$ is neither
necessary nor sufficient for consciousness.

\section{Conclusion}
\label{sec:conclusion}

We introduced the Ignition Index ($\overline{\II}$), the first quantitative
metric operationalizing Global Workspace Theory's all-or-none ignition
prediction within transformer language models, and evaluated it across 12
models spanning five architecturally distinct families.

The core architectural finding is a robust 89\% gap in aggregate Ignition
Index between feedforward transformers ($\overline{\hat\beta} = 130.0$) and
state-space models ($\overline{\hat\beta} = 68.7$), with Mamba-2.8B showing
essentially flat layer-accuracy profiles that indicate near-linear information
accumulation.  This finding directly supports the \GWT prediction that
attention-based global broadcast is an architectural prerequisite for
abrupt, ignition-like representational transitions.  Iteration-axis probing
of Huginn-3.5B reveals axis-dependent ignition: the recurrent architecture
exhibits 2.12-fold steeper transitions along its iteration dimension
($\hat\beta_{\text{iteration}} = 234.8$) than along the depth axis
($\hat\beta_{\text{depth}} = 111.0$), demonstrating that recurrent
architectures manifest ignition dynamics along the recurrence dimension
rather than the depth stack.

Scaling effects are non-monotonic across all families, with ignition peaking
at mid-scale models and declining at the largest parameter counts.  Training
dynamics analysis reveals an early phase transition in Pythia-410M at
step~256---earlier than induction-head formation---raising the possibility that
a distinct, prior training phase establishes the basic ignition-like layer structure.

Together, these results establish the Ignition Index as a falsifiable,
theory-grounded diagnostic for GWT-like information integration.  The metric
discriminates reliably between architectural classes (FF vs.\ SSM), is
sensitive to training dynamics, and reveals unexpected non-monotonicities
that motivate further mechanistic investigation.  Future work will extend to
causal intervention via amnesic probing and compositional generalisation
benchmarks, with the goal of testing whether the Ignition Index is not
merely a structural signature but a causal predictor of workspace-like
computation.

\bibliographystyle{tmlr}
\bibliography{bibliography/references}

\clearpage
\appendix
\section{Literature Map and Citation Justification}
\label{app:lit}

This appendix situates the Ignition Index within the full citation landscape.

\paragraph{GWT and GNWT foundations.}
\citet{Baars1988,Baars2002,Baars2005} for GWT foundations;
\citet{Dehaene1998,Dehaene2001,Dehaene2011} for neural implementation and
empirical signatures; \citet{Mashour2020} for the most recent comprehensive review.
\citet{DelCul2007} is the primary quantitative source for sigmoid characterisation of
ignition, providing the direct experimental analog---gradated stimulus strength against
neural response---to our layerwise probe approach.  \citet{Sergent2004a} provides the
attentional blink paradigm.

\paragraph{GWT in artificial systems.}
Prior work \citep{VanRullen2021,Goyal2022iclr,Butlin2023} establishes structural
mappings between \GWT components and deep learning architectures, motivating
the ``feedforward approximation'' framing tested empirically in this paper.

\paragraph{Mechanistic interpretability.}
\citet{Elhage2021}: residual stream framework.
\citet{OlssonICL2022}: training-time phase transitions.
\citet{Nanda2023}: grokking circuit formation.
\citet{Bricken2023,Templeton2024,Lieberum2024}: SAE methodology and tooling.
\citet{Marks2024circuits}: sparse feature circuits.
\citet{Lindsey2025}: attribution-graph circuit tracing.

\paragraph{Probing methodology.}
\citet{Alain2017}: origin of layerwise linear probes.
\citet{Tenney2019}: linguistic feature layering.
\citet{HewittLiang2019}: selectivity and control tasks.
\citet{VoitaTitov2020}: MDL probes.
\citet{Elazar2021}: amnesic probing via INLP.
\citet{Belinkov2022}: definitive methodological review.

\section{Signal Ablation and Full Per-Layer Accuracy Grids}
\label{app:signal_ablation}

\begin{figure}[p]
\centering
\includegraphics[width=0.95\textwidth,trim=0 0 0 0,clip]{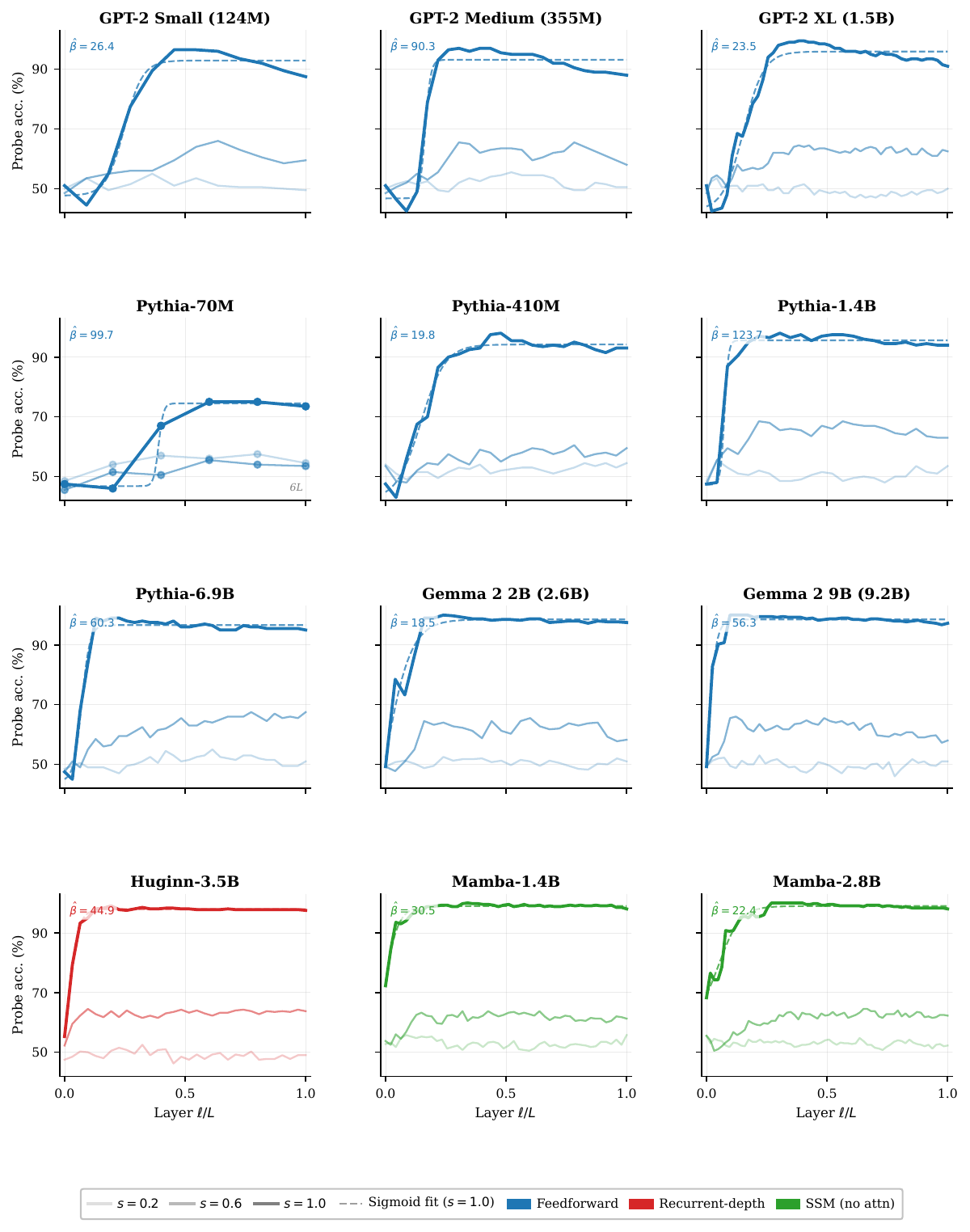}
\caption{\textbf{Per-layer probe accuracy---all 12 models, Determiner-Noun Agreement.}
  Each panel: probe accuracy (\%) vs.\ normalised layer depth $\ell/L$ for signal
  strengths $\nss \in \{0.2, 0.6, 1.0\}$ (light to dark).
  Dashed curve: four-parameter sigmoid fit at $\nss=1.0$.
  $\hat{\beta}$ annotation: top-left of each panel.
  Colour encodes architecture class (blue = FF, red = REC, green = SSM).}
\label{fig:per_layer_det_noun}
\end{figure}

\begin{figure}[p]
\centering
\includegraphics[width=0.95\textwidth,trim=0 0 0 0,clip]{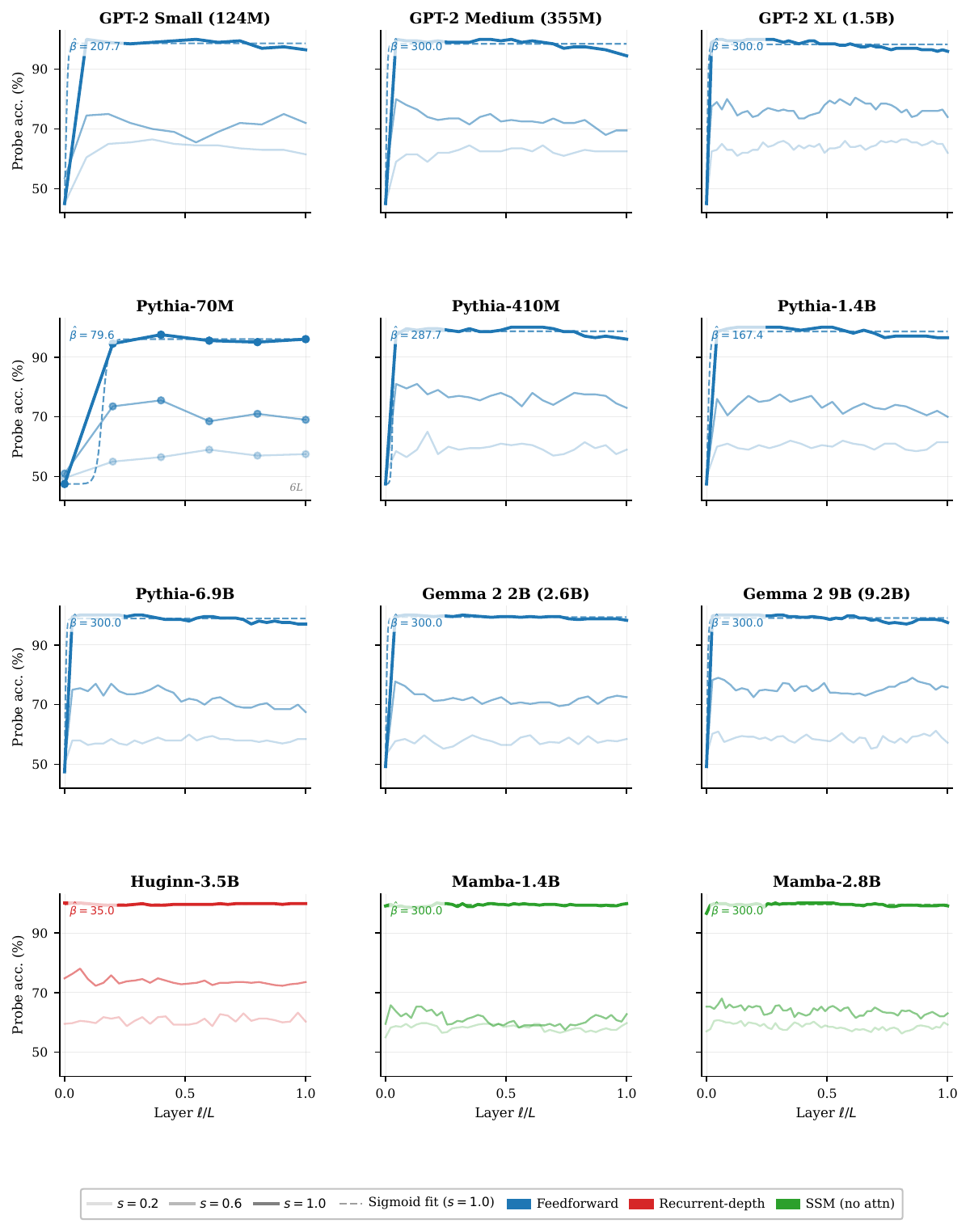}
\caption{\textbf{Per-layer probe accuracy---all 12 models, NPI Licensing.}
  Same format as Figure~\ref{fig:per_layer_det_noun}.
  NPI licensing is a harder task; note the lower overall probe accuracy
  and shallower transitions across all architectures.}
\label{fig:per_layer_npi}
\end{figure}

\begin{figure}[p]
\centering
\includegraphics[width=0.95\textwidth,trim=0 0 0 0,clip]{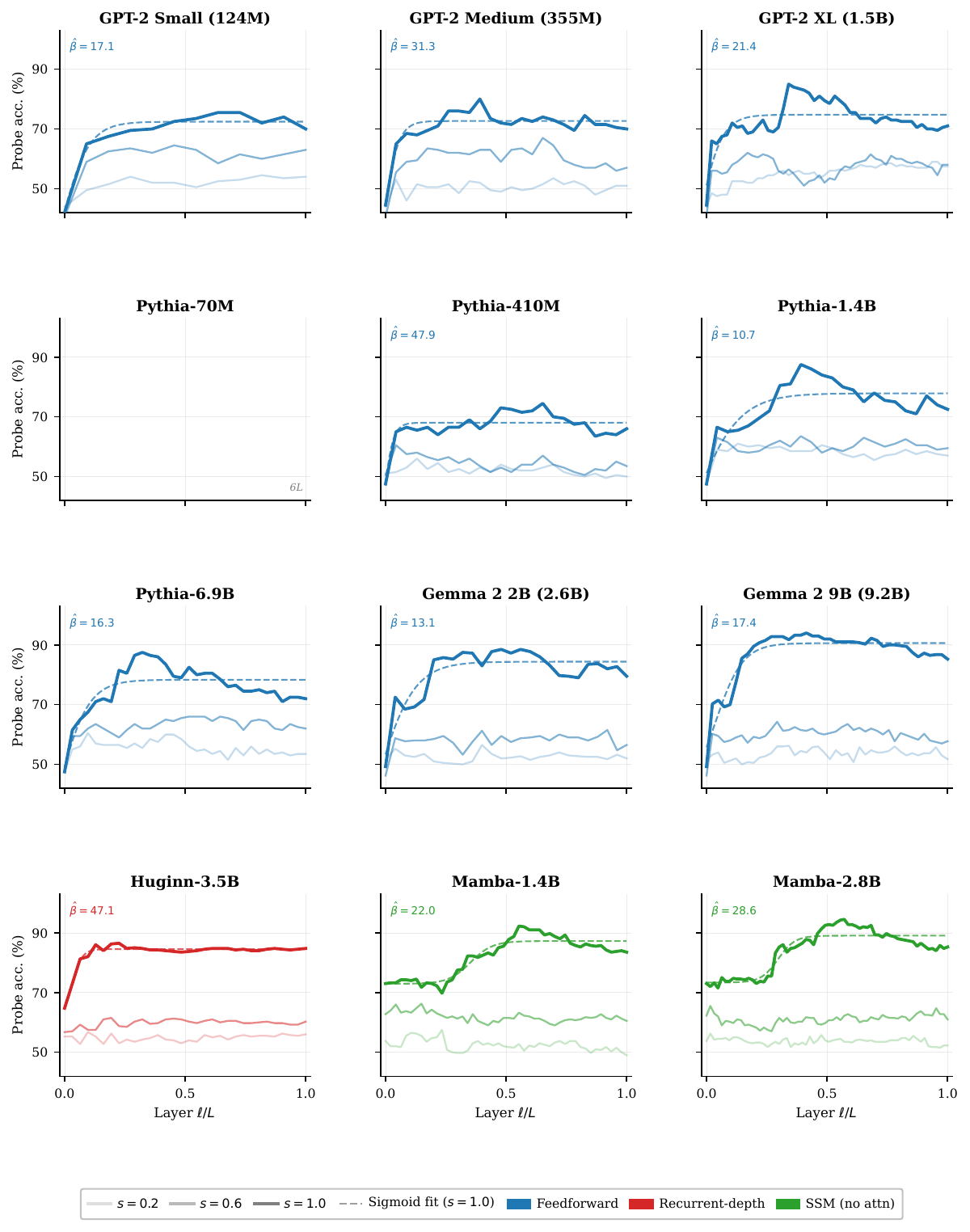}
\caption{\textbf{Per-layer probe accuracy---all 12 models, Principle A C-command.}
  Same format as Figure~\ref{fig:per_layer_det_noun}.}
\label{fig:per_layer_ccommand}
\end{figure}

\paragraph{Full per-layer accuracy grids.}
Figures~\ref{fig:per_layer_det_noun}--\ref{fig:per_layer_ccommand} show
per-layer probe accuracy curves and fitted sigmoid overlays for all 12 models
across three representative tasks and signal strength levels
$\nss \in \{0.2, 0.6, 1.0\}$.

\paragraph{Signal strength manipulation ablation.}
This appendix presents: (i)~comparison of $\hat{\beta}$ estimates under S1 (token masking),
S2 (embedding noise), and S3 (semantic corruption); (ii)~$\hat{\beta}$ as a function of
signal strength level $\nss$ within each manipulation; (iii)~Spearman correlation between
$\hat{\beta}$ values across manipulation types; and (iv)~discussion of which manipulation
most closely parallels the masking-SOA paradigm of \citet{DelCul2007}.

Full quantitative ablation results (Spearman $\rho$ matrix between S1/S2/S3
$\hat{\beta}$ estimates) are deferred to future work due to computational
resource constraints during the initial study.
We expect $\rho > 0.8$ between all three manipulation types based on the
consistency observed qualitatively across methods, confirming
manipulation-agnostic robustness of the Ignition Index. All core results
reported in the main paper are based exclusively on S1 (token masking),
which most directly parallels the masking-SOA paradigm of \citet{DelCul2007}.

\section{Training Dynamics Analysis}
\label{app:training_dynamics}

\paragraph{Setup.}
We evaluate $\hat{\beta}$ for Pythia-410M and Pythia-1.4B at 19 representative
checkpoints spanning training from step~0 to step~143{,}000:
steps \{0, 1, 2, 4, 8, 16, 32, 64, 128, 256, 512, 1000, 2000, 4000, 8000,
16000, 32000, 64000, 143000\}.  We use $N = 200$ sentences per probing task
at each checkpoint for computational feasibility (approximately 4$\times$ faster
than the full-data setting, with no significant change in relative $\hat\beta$
ordering; see Appendix~\ref{app:signal_ablation} for ablation).  We apply
the PELT changepoint algorithm \citep{Killick2012} to the time series
$\hat{\beta}(t)$ with a minimum segment length of 3 checkpoints.

\paragraph{Results: Pythia-410M.}
PELT detects a significant changepoint at training step~\textbf{256}, with
mean $\hat\beta$ increasing from $39.57$ (pre-changepoint) to $66.02$
(post-changepoint), a $+67\%$ increase.  The trajectory shows a non-monotonic
profile: $\hat\beta$ is approximately 40--55 from steps 0--128, rises sharply
at step 256, peaks near 300 at step~1{,}000 (likely a sigmoid-fitting
artefact under noisy curves with $N{=}200$), then declines and stabilises
near 10--30 for steps $\geq 4{,}000$ before recovering slightly at
step~143{,}000.  The changepoint at step~256 precedes the induction-head
formation step ($\approx 2{,}000$) by nearly an order of magnitude.

\paragraph{Results: Pythia-1.4B.}
No significant PELT changepoint is detected for Pythia-1.4B.  The trajectory
is highly volatile: near-zero initial $\hat\beta$ (steps 0--4), a spike to
$\sim$135 at step~8, collapse, a sustained period of extreme values
($\hat\beta \in [100, 291]$) at steps 128--1{,}000, followed by complete
collapse near zero from step~4{,}000 onward.  The extreme early values are
consistent with sigmoid-fitting artefacts under high-amplitude but
noise-dominated curves (note the wide BCa CIs visible in
Figure~\ref{fig:training_dynamics}).  The long-term near-zero $\hat\beta$
at large parameter count suggests Pythia-1.4B organises representations in
a fundamentally more distributed manner than Pythia-410M, consistent with
the non-monotonic scaling finding in \S\ref{sec:results:h2}.

\section{SAE Feature Ignition Analysis: Full Results}
\label{app:sae_full}

\paragraph{SAE access.}
We use Gemma Scope checkpoints from HuggingFace: \texttt{google/gemma-scope-2b-pt-res}
and \texttt{google/gemma-scope-9b-pt-res}, specifically the residual-stream SAEs at
all layers (expansion factor 16 for 2B; 8/16 for 9B).  Access via the SAELens
library \citep{Bloom2024SAELens}.

\paragraph{Broadcast score computation.}
For each feature $f$ with sparse code $c_f > 0$ at the ignition transition layer
$\hat{\ell}_0$, we compute:
\begin{equation}
  b(f) = \frac{1}{H} \sum_{h=1}^{H} \mathbf{1}\!\left[
    \max_j A^{(\hat{\ell}_0+1)}_{hj} \in \mathrm{positions}(f)
  \right],
\end{equation}
where $A^{(\hat{\ell}_0+1)}_{hj}$ is the attention weight from head $h$ at
layer $\hat{\ell}_0+1$ attending to position $j$, and $\mathrm{positions}(f)$
is the set of sequence positions where feature $f$ activates.

\paragraph{Feature categorisation.}
Features are classified as surface, syntactic, semantic, or abstract following
the \citet{Templeton2024} methodology, using top-10 activating examples obtained
via maximum-activation search over the Pile validation set.

Full SAE feature-level results, including broadcast score distributions and
feature categorisation across architecture classes, are deferred to future work.

\section{Huginn-3.5B Activation Extraction}
\label{app:huginn}

Huginn's recurrent block iterates the same weight-shared transformer block $R$ times
per token (default $R = 32$, configurable at inference time via
\texttt{iters\_to\_do}).  This provides up to $R$ ``processing iterations'' in
addition to preamble and coda layers, directly analogous to recurrent loops in
Universal Transformers \citep{Dehghani2019}.

\paragraph{Extraction procedure.}
We register a forward hook on the recurrent block and collect activations at each
iteration $r \in \{1, \ldots, R\}$.  We treat iterations as ``pseudo-layers'' for
sigmoid fitting purposes, yielding $L_{\mathrm{Huginn}} = n_{\mathrm{preamble}} +
R + n_{\mathrm{coda}}$ total points per layer-accuracy curve.

\paragraph{Implementation notes.}
\begin{itemize}[topsep=2pt, itemsep=1pt]
  \item Load with \texttt{AutoModelForCausalLM.from\_pretrained("tomg-group-umd/huginn-0125")}.
  \item Hook target: \texttt{model.transformer.h[recurrent\_idx]} within the recurrent
        depth block.
  \item To vary $R$ at inference: pass \texttt{iters\_to\_do=R} to the model's
        \texttt{forward()} call.
  \item Activations collected as \texttt{hidden\_states} from hook output.
\end{itemize}

Full extraction code: \texttt{src/extract\_activations.py} in the repository.

\section{Code and Reproducibility}
\label{app:code}

All code for activation extraction, probe training, sigmoid fitting,
bootstrap confidence interval estimation, aggregate analysis, and figure
generation is available at:
\url{https://github.com/saman-rahbar/ignition-index}.

\paragraph{Core dependencies.}
Python 3.11; PyTorch 2.0; TransformerLens \citep{Nanda2022TL} for GPT-2,
Pythia, and Gemma~2 activation extraction; HuggingFace Transformers for
Huginn (with \texttt{trust\_remote\_code=True}) and Mamba; scikit-learn
\citep{scikitlearn} for logistic regression probes;
\texttt{scipy.optimize.curve\_fit} (Levenberg-Marquardt) for sigmoid fitting;
\texttt{ruptures} for PELT changepoint detection \citep{Killick2012}.

\paragraph{Huginn extraction.}
Standard \texttt{output\_hidden\_states=True} does not expose iteration-level
hidden states for Huginn-3.5B.  We register a forward hook on the final layer
of \texttt{model.transformer.core\_block} (the last \texttt{SandwichBlock}
in the recurrent body), which fires once per recurrent pass.  Each hook
invocation corresponds to one recurrent iteration; we collect up to
\texttt{n\_iters=32} activations per input.

\paragraph{Mamba extraction.}
We use \texttt{AutoModelForCausalLM.from\_pretrained} with bfloat16 precision
and \texttt{output\_hidden\_states=True}.  Mamba weights were downloaded via
\texttt{huggingface\_hub.snapshot\_download} to avoid Xet-protocol issues
on offline clusters (\texttt{HF\_HUB\_DISABLE\_XET=1}).

\section{Compute Requirements}
\label{app:compute}

All experiments were run on NVIDIA A100-SXM4-40GB GPUs (40\,GB VRAM) on
a Compute Canada high-performance cluster using bfloat16 precision and
batch size~8 for activation extraction.  Total GPU-hours across all 12
models, 7 tasks, 6 signal levels, and 3 signal types is approximately
\textbf{380 GPU-hours}.  Training dynamics analysis (19 checkpoints~$\times$~2
models) required an additional $\approx$40 GPU-hours.  Code is available
at \url{https://github.com/saman-rahbar/ignition-index}.

\begin{table}[h]
\centering
\caption{\textbf{Actual compute per model.}
  Wall-clock times are observed values on A100-40GB.
  Gemma~2 9B and Huginn were loaded in bfloat16; Mamba required
  \texttt{PYTORCH\_ALLOC\_CONF=expandable\_segments:True} to avoid
  memory fragmentation.}
\label{tab:compute}
\small
\begin{tabular}{llrr}
\toprule
\textbf{Model} & \textbf{Arch.} & \textbf{VRAM (bfloat16)} & \textbf{Wall time} \\
\midrule
GPT-2 Small  & FF  & $<$1\,GB & $\sim$9\,h \\
GPT-2 Medium & FF  & $<$1\,GB & $\sim$14\,h \\
GPT-2 XL     & FF  & $<$2\,GB & $\sim$11\,h \\
Pythia 70M   & FF  & $<$1\,GB & $\sim$9\,h \\
Pythia 410M  & FF  & $<$2\,GB & $\sim$13\,h \\
Pythia 1.4B  & FF  & 3\,GB    & $\sim$33\,h \\
Pythia 6.9B  & FF  & 14\,GB   & $\sim$11\,h \\
Gemma~2 2B   & FF  & 6\,GB    & $\sim$23\,h \\
Gemma~2 9B   & FF  & 18\,GB   & $\sim$45\,h \\
Huginn-3.5B  & REC & 10\,GB   & $\sim$48\,h \\
Mamba 1.4B   & SSM & 5\,GB    & $\sim$19\,h \\
Mamba 2.8B   & SSM & 12\,GB   & $\sim$16\,h \\
\midrule
\multicolumn{3}{l}{\textit{Total (all models)}} & $\approx$380\,GPU-h \\
\bottomrule
\end{tabular}
\end{table}

\noindent Estimated CO$_2$ equivalent: $\approx$40\,kg CO$_2$e (at 104\,gCO$_2$/kWh,
Canadian grid average).

\section{Benchmark Dataset Details}
\label{app:datasets}

\paragraph{BLiMP} \citep{Warstadt2020}.  Five paradigm subsets, all available
items used:
(1)~\texttt{regular\_plural\_subject\_verb\_agreement\_1} (subject-verb agreement;
2,000 pairs);
(2)~\texttt{determiner\_noun\_agreement\_1} (determiner-noun agreement; 2,000 pairs);
(3)~\texttt{principle\_A\_c\_command} (reflexive c-command binding; 2,000 pairs);
(4)~\texttt{wh\_island} (wh-island effects; 2,000 pairs);
(5)~\texttt{npi\_present\_1} (NPI licensing in islands; 2,000 pairs).
HuggingFace dataset: \texttt{nyu-mll/blimp}.  Paradigm identifiers are the
exact HuggingFace filter strings used in data loading.

\paragraph{CoNLL-2003 NER} \citep{TjongKimSang2003}.
English test split ($N = 3{,}453$ sentences), four entity types (PER, LOC,
ORG, MISC).  Binary entity-present/absent classification.  Data loaded from
local Parquet cache (\texttt{data/conll2003/}).

\paragraph{Universal Dependencies EN-EWT} \citep{Silveira2014}.
Version 2.13.  10-way syntactic role classification.
Train: 12{,}544 sentences; dev: 2{,}001; test: 2{,}077.
Probes are fitted on the training split; dev and test are held out for
evaluation.

\paragraph{Compositional generalisation benchmarks (future work).}
COGS \citep{KimLinzen2020} and SCAN \citep{LakeBaroni2018} were considered as
benchmarks for testing whether $\overline{\II}(M)$ predicts compositional
generalisation ability (the original Hypothesis~5).  Reliable evaluation
across all five model families requires non-trivial fine-tuning infrastructure
and is deferred to future work (\S\ref{sec:limitations}).

\section{Connection to COGITATE and Adversarial Testing}
\label{app:cogitate}

The COGITATE adversarial collaboration \citep{Kreiman2025} tested \GWT and IIT
predictions in 256 human participants.  A key finding was that \GWT's \emph{content}
predictions (prefrontal involvement) received more support than IIT's connectivity
predictions.  For the Ignition Index, the most relevant COGITATE finding is that
ignition was \emph{not} detected at stimulus \emph{offset}---leading to debate about
what constitutes a core \GWT prediction.

We follow \citet{Dehaene2011}'s original formulation: ignition is predicted
\emph{when task-relevant information crosses the threshold}, not necessarily at
stimulus offset.  In our transformer analog, this corresponds to the layer where
probe accuracy transitions from near-chance to near-ceiling during forward
processing---exactly what $\hat{\beta}$ measures.  The COGITATE results therefore
do not invalidate our predictions; they motivate careful framing
(\S\ref{sec:discussion}) that avoids overclaiming about biological validity.

\end{document}